\pdfoutput=1

\documentclass[11pt]{article}

\usepackage[preprint]{acl}

\usepackage{times}
\usepackage{latexsym}

\usepackage[T1]{fontenc}

\usepackage[utf8]{inputenc}

\usepackage{microtype}

\usepackage{inconsolata}

\usepackage{graphicx}
\usepackage{colortbl}
\usepackage{multirow}
\usepackage{amsmath}

\title{CEAID: Benchmark of Multilingual Machine-Generated Text Detection Methods for Central European Languages}

\author{Dominik Macko \and Jakub Kopal\\
  Kempelen Institute of Intelligent Technologies \\
  \texttt{dominik.macko@kinit.sk}, \texttt{jakub.kopal@kinit.sk} \\}

\begin{document}
\maketitle
\begin{abstract}
Machine-generated text detection, as an important task, is predominantly focused on English in research. This makes the existing detectors almost unusable for non-English languages, relying purely on cross-lingual transferability. There exist only a few works focused on any of Central European languages, leaving the transferability towards these languages rather unexplored. We fill this gap by providing the first benchmark of detection methods focused on this region, while also providing comparison of train-languages combinations to identify the best performing ones. We focus on multi-domain, multi-generator, and multilingual evaluation, pinpointing the differences of individual aspects, as well as adversarial robustness of detection methods. Supervised finetuned detectors in the Central European languages are found the most performant in these languages as well as the most resistant against obfuscation.
\end{abstract}

\section{Introduction}
\label{sec:intro}

Large language models (LLMs) are able to generate texts in various languages, hardly distinguishable for humans from authentic human-written texts. However, automated detection of such texts is mostly researched for English only (or other high-resource languages, such as Spanish or Russian), leaving some languages unprotected from massive spread of AI-generated content for malicious purposes (e.g., disinformation, spam, frauds, plagiarism). Such languages are often left to rely on cross-lingual transfer of monolingual detectors, which can have severely degraded performance.

\begin{figure}[!t]
\centering
\includegraphics[width=0.8\linewidth]{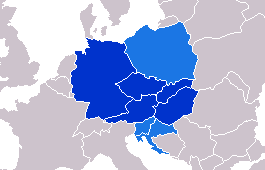}
\caption{Central European region as defined by \citealp{bideleux2007history}.}
\label{fig:region}
\raggedright \small source:{\scriptsize\url{https://en.wikipedia.org/wiki/Central\_Europe#/media/File:Central-Europe-SwanseaUniv.png}}
\vspace{-3mm}
\end{figure}

It is important to explore the effect of such cross-lingual transferability to such languages and possibilities of introduction of multilingual detectors with an involvement of these languages in the detectors training or finetuning process. To the best of our knowledge, there is no study available focused on machine-generated text (MGT) detection in the languages of Central European region (see \figurename~\ref{fig:region}), especially focused on cross-lingual transferability.

Our study is specifically focused on answering the following research questions.
\textit{\textbf{RQ1:} Are there differences in finetuned detection methods performance based on combination of train languages?} If so, which combination of train languages makes the detectors the most generalizable to the other languages of the Central European region?
\textit{\textbf{RQ2:} Which category of detection methods is most suitable for Central European languages?} Are there differences among the MGT generation models?
\textit{\textbf{RQ3:} Which detection methods are most robust against obfuscation in Central European languages?} Are there differences in such robustness among the selected languages?

The contributions of our work can be summarized as\footnote{For replication possibilities, all source code and data will be released for non-commercial research purpose upon paper acceptance.}:
\begin{itemize}
    \item \textbf{the first comprehensive} (multilingual, multi-domain, multi-generator) \textbf{benchmark of MGT} detection methods focused on \textbf{Central European region}, reflecting cultural context of this specific region,
    \item the evaluation and \textbf{comparison of differences} in performance of MGT detection methods between \textbf{news and social-media domains}, reflecting different lengths and styles (e.g., formality, grammar, emoticons) of texts of these two domains,
    \item the evaluation of \textbf{train language combination effect} on generalizability and adversarial robustness of MGT detection methods, identifying the most suitable combination of train languages for the Central European region,
    \item introduction of a bunch of \textbf{language-specific MGT detection methods} for under-researched languages, which are usually out-of-focus of the mainstream MGT research relying purely on cross-lingual transferability of detectors (i.e., degraded performance).
\end{itemize}

This paper is structured in the following way. Firstly, we provide an overview of existing works related to the MGT detection in some of the selected Central European languages. In Section~\ref{sec:methodology}, we describe the used dataset and methodology for the experiments. Then, we provide the experimental results along with a discussion in Section~\ref{sec:results}. At the end, we summarize the conclusions of our study.

\section{Related Work}
\label{sec:related}

The related works are reviewed in three groups. First, the studies of applying MGT detection to non-English languages are summarized with interesting observations. Second, the existing MGT detection shared tasks are overviewed that are focusing not purely on English. Lastly, the existing benchmark datasets are summarized that focus on multiple languages, especially those that could be utilized in cross-lingual study.

\subsection{Non-English MGT Detection}
A study covering four languages (English, French, German, Spanish) of (South-)West European region shows that statistical features primarily developed for English MGT detection can be used in other languages as well \citep{schaaff-etal-2023-classification}. The detection has been however realized in monolingual way, i.e. testing on a single language that was included in training. Similarly, a study~\citep{uyuk-etal-2024-crafting}, focused on English, Turkish, Hungarian, and Persian, has used a machine-learning classifiers on top of the TF-IDF features for monolingual detection.
Another study, focused on academic integrity, tested multiple existing directly usable (i.e., without training) detectors to examine their performance in machine translated texts \citep{weber2023testing}. The human written texts in Bosnian, Czech, German, Latvian, Slovak, Spanish, and Swedish, have been translated to English for the detection. They revealed that machine translated text had higher false positive rate than the text written by humans directly in English. It means that machine translation is not usable for MGT detection, and detection directly in non-English languages must be used. This is further supported by similar studies \citep{vsigut2023can, bohacek2023unseen}, where the authors tested detection directly in Czech and Slovak and compared with translation to English. The translation negatively affected the performance of the used multilingual detector. They further identified that ChatGPT-4 was harder to detect than ChatGPT-3.5, pinpointing the need to conduct evaluation studies with most modern set of generators.
Most of the above mentioned studies have used ChatGPT (or its variants) as the only MGT generator (\citealp{uyuk-etal-2024-crafting} used 5 different LLMs). Therefore, the generalizability of their conclusions is questionable.
The study on detection of generated German texts \citep{irrgang-etal-2024-features} pin-pointed a worse cross-generator transferability of the detection performance, requiring robust detection methods.
The study on detection of LLM-generated emails in Polish \citep{10.1145/3664476.3670465} has identified that detection in underrepresented language in models training is inferior.
The study on Bulgarian social-media texts \citep{temnikova-etal-2023-looking} has shown that finetuned detectors on texts in that particular language can boost the performance significantly.

\subsection{MGT Detection Shared Tasks}
To drive the research direction into particular languages, there have recently been multiple MGT detection shared tasks focused monolingually on Russian at RuATD~2022~\citep{Shamardina_2022}, Spanish at AuTexTification~2023~\citep{sarvazyan2023overview}, Dutch at CLIN33~\citep{clin33}. The IberAuTexTification~2024~\citep{sarvazyan2024overview} shared task was focused on multidomain detection in six languages (English, Spanish, Portuguese, Catalan, Basque, and Galician), especially targeting Iberian peninsula. It represents a regional focus for development of MGT detection methods, especially for low-resource languages; however, the detection in these languages has been executed individually (i.e., also in monolingual way). Thus, cross-lingual transferability aspects have not been examined.
Recent multilingual shared tasks include SemEval-2024 Task~8~\citep{wang-etal-2024-semeval-2024} and GenAI Content Detection Task~1~\citep{wang-etal-2025-genai}, both of them representing mutli-generator, multi-domain, and multilingual MGT detection challenge. However, in both of them, there is inconsistency in coverage of combinations of generators and domains across languages; thus any comparison among performances between languages is inherently biased. Furthermore, cross-lingual aspects are evaluated only towards few unseen languages during training; leaving the effect of cross-lingual transfer from particular languages unexplored.

\subsection{Multilingual MGT Detection Benchmarks}
The MultiSocial~\citep{macko-etal-2025-multisocial} benchmark is focused on comparison of performance of MGT detection methods for social-media texts of 5 platforms in 22 languages. For generation of MGT is uses 7 modern LLMs that 3-times paraphrase the original human-written texts. The number of samples per each platform and per each language is however not consistent, making the cross-lingual transferability experiments rather limited. The M4GT-Bench~\citep{wang-etal-2024-m4gt} dataset has been used in the above-mentioned multilingual shared tasks. Inconsistency between per-language domain and generation settings makes the cross-lingual experiments inherently biased. Another benchmark focused on multilingual news articles, called MULTITuDE~\citep{macko-etal-2023-multitude}, covers 11 languages; however, only 3 of them contain training samples (English, Russian, and Spanish). The authors focus on cross-lingual transferability of detectors trained on these three languages towards the others in the test set. It is further extended to evaluation of adversarial robustness against 10 authorship obfuscation methods~\citep{macko-etal-2024-authorship}. It has identified a homoglyph-based obfuscation especially successful to evade the detection in multilingual settings. The RAID benchmark~\citep{dugan-etal-2024-raid}, including RAID-extra part containing besides English also Python code and news articles in German and Czech, is focused on variable generation setting (decoding strategy) and 11 adversarial attacks. Such data are suited for evaluation of detectors robustness; however, language limitation disqualifies it from cross-lingual experiments.

\section{Methodology}\vspace{-1mm}
\label{sec:methodology}

In order to provide the answers for our research questions, stated in Introduction, we need to craft a suitable dataset for the experiments with a good selection of languages, select suitable multilingual MGT detection methods of different categories for comparison, and come up with a proper settings for rigorous evaluation enabling generalizability of the conclusions. All of these are reflected in the following subsections, while limiting the scope of the study to keep it feasible and prevent waste of resources (especially time and computational requirements).

\subsection{Dataset}\vspace{-1mm}

\begin{table}[!b]
\centering
\resizebox{\linewidth}{!}{
\begin{tabular}{l|cc|cc|cc}
\hline
\textbf{Domains $\rightarrow$} & \multicolumn{2}{c|}{\bfseries News} & \multicolumn{2}{c|}{\bfseries Social media} & \multicolumn{2}{c}{\bfseries All} \\
\textbf{Language $\downarrow$} & \textbf{Train} & \textbf{Test} & \textbf{Train} & \textbf{Test} & \textbf{Train} & \textbf{Test} \\
\hline
cs (Czech) &        7734 &       2328 &         11041 &         6073 &  18775 &  8401 \\
de (German) &        7764 &       2322 &         21038 &         9497 &  28802 & 11819 \\
hr (Croatian) &        7819 &       2348 &         14475 &         5993 &  22294 &  8341 \\
hu (Hungarian) &        7791 &       2350 &         14492 &         5957 &  22283 &  8307 \\
pl (Polish) &        7818 &       2336 &         16687 &         6971 &  24505 &  9307 \\
sk (Slovak) &        7664 &       2317 &             0 &         2026 &   7664 &  4343 \\
sl (Slovenian) &        7845 &       2354 &             0 &         3058 &   7845 &  5412 \\
\hline
\textbf{Total} &       54435 &      16355 &         77733 &        39575 & 132168 & 55930 \\
\hline
\end{tabular}
}
\vspace{-2mm}
\caption{Overview of the selected dataset sample counts.}
\label{tab:dataset}
\vspace{-2mm}
\end{table}

Our goal was to ensure consistency among languages, enabling rigorous cross-lingual evaluation. For generalization of our observations, we included multiple generators and multiple domains. Specifically, we have selected the Central European languages from MULTITuDE\_v3~\citep{macko_2025_15519413} (domain of news articles) and
MultiSocial~\citep{macko-etal-2025-multisocial} (domain of social-media texts from 5 platforms), having at least 200 samples (for test split) per each domain and class (human vs. machine). It resulted into selection of 7 languages, namely Croatian, Czech, German, Hungarian, Polish, Slovak, and Slovenian, together covering 3 language-family branches of Germanic, Slavic, and Uralic (all using Latin writing script). 2 languages, Slovak and Slovenian, are not used for training due to not having enough samples from the social-media domain available. The overview of the resulted sample counts per each language, domain, and split of the selected dataset is provided in Table~\ref{tab:dataset}. The MGTs are generated by 8 LLMs in total, 6 of which are the same across the two domains (Aya-101, GPT-3.5-Turbo-0125, Mistral-7B-Instruct-v0.2, OPT-IML-Max-30B, v5-Eagle-7B-HF, and Vicuna-13B), one is only in news domain (Llama-2-70B-chat-hf), and one is only in social-media domain (Gemini). Since the generators are consistent among the languages (uniformly distributed sample counts), we do not see a problem to include all of them in regard to our research questions (not only the intersection).

\subsection{Detectors}

We follow the previous cross-lingual benchmark studies~\citep{macko-etal-2023-multitude, macko-etal-2025-multisocial} to select 3 categories of MGT detection methods: \textit{statistical} (zero-shot), pretrained (directly applicable), and finetuned (trained on train split).

As statistical detectors, we are using \textbf{Binoculars}~\citep{10.5555/3692070.3692768}, \textbf{Fast-DetectGPT}~\citep{bao2023fast}, and \textbf{LLM-Deviation}~\citep{wu2023mfd}, all of them based on multilingual mGTP LLM~\citep{shliazhko-etal-2024-mgpt}.

As pretrained detectors, we have selected \textbf{ChatGPT-detector-RoBERTa-Chinese}~\citep{guo-etal-2023-hc3}, \textbf{Longformer Detector}~\citep{li-etal-2024-mage}, \textbf{BLOOMZ-3B-mixed-detector}~\citep{sivesind_2023}. All of the mentioned detectors have been selected due to performing well in each category in the MultiSocial~\citep{macko-etal-2025-multisocial} study or its cross-domain evaluation. Analogously to the above mentioned study, we have used the published source code of the IMGTB framework~\citep{spiegel-macko-2024-imgtb} to run these detectors.

As finetuned detectors, we have selected \textbf{mDeBERTa-v3-base}~\citep{he2021debertav3} and 
\textbf{XLM-RoBERTa-base}~\citep{DBLP:journals/corr/abs-1911-02116} (as multilingual baselines), \textbf{Llama-3.2-3B}~\citep{meta2024llama} (as a newer and smaller version of the best performing MultiSocial-finetuned model), and \textbf{Gemma-2-2B}~\citep{gemma_2024} (as a smaller version of the best performing model in out-of-distribution evaluation of \citealp{macko2025increasingrobustnessfinetunedmultilingual}).

\subsection{Settings}

For strong representativeness of our results and generalization of conclusions, we have carefully designed the experiments. Targeting RQ1, we have selected 4 models for finetuning of diverse architectures (encoder vs decoder) and sizes (number of parameters ranging from 0.3B to 3B). For RQ2, we have used 3 categories of detection methods, each containing at least 3 methods (based on existing studies). For RQ3, we have used two prominent obfuscation methods (paraphrasing for its usability and homoglyphs for their effectiveness) and 10 diverse detectors.

As the primary evaluation metric, we use \textbf{AUC ROC} (area under curve of the receiver operating characteristic) as a classification-threshold independent metric. We are also providing \textbf{TPR @ 5\% FPR} in the appendix for a deeper analysis, representing true positive rate (TPR) using the classification thresholds calibrated (based on ROC curve) to reach 5\% of false positive rate (FPR), reflecting an expected performance in the wild.

To further ensure consistency between the two domains and between the two classes (human and machine), in each experiment, we have used pseudo-random sub-sampling of training samples to the highest possible number to achieve the perfect balance. This number is the lowest count of human-written texts out of each domain for each language, being 986. It resulted into training set of 3944 samples per language ($986 \times 2~\text{[human vs machine]} \times 2~\text{[news vs social-media]}$). This number of train samples is kept the same in all experiments (monolingual training as well as multilingual training with even portion of each train language). For testing and unbiased comparison, we have further sub-sampled 250 samples from the test split per each class, domain, and language, and 200 samples per each generator, domain and language for comparison of detection performance per generators. These numbers are reflecting the lowest count of test samples for any combination to achieve perfect balance.

To save computational resources for adversarial robustness evaluation, we have sub-sampled 100 samples per each class, domain, and language from test set to be further paraphrased by DeepSeek-R1-Distill-Qwen-32B~\citep{deepseekai2025deepseekr1incentivizingreasoningcapability} (as a most modern highly-performant multilingual LLM) and obfuscated by the generic HomoglyphAttack~\citep{macko-etal-2024-authorship}. The comparison between above-mentioned (bigger) test set evaluation and this smaller subset ensures its representativeness. From the finetuned models (31 per each of 4 base models), we have selected for evaluation of adversarial robustness only de-hu-pl trained versions of each base model (as representatives of the three different language-family branches and achieving one of the highest performances).

\section{Results}
\label{sec:results}

The results are divided into three parts based on the addressed research questions.

\begin{table*}[!t]
\centering
\resizebox{0.95\textwidth}{!}{
\addtolength{\tabcolsep}{-2pt}
\begin{tabular}{l|cc|cc|cc|cc|cc|cc|cc|cc}
\hline
\textbf{Train} & \multicolumn{2}{c|}{\textbf{All}} & \multicolumn{2}{c|}{\textbf{cs}} & \multicolumn{2}{c|}{\textbf{de}} & \multicolumn{2}{c|}{\textbf{hr}} & \multicolumn{2}{c|}{\textbf{hu}} & \multicolumn{2}{c|}{\textbf{pl}} & \multicolumn{2}{c|}{\textbf{sk}} & \multicolumn{2}{c}{\textbf{sl}} \\
\cline{2-17}
\textbf{Languages} & mean & std & mean & std & mean & std & mean & std & mean & std & mean & std & mean & std & mean & std \\
\hline
cs-de-hr-hu-pl & \bfseries 0.967 & 0.005 & 0.982 & 0.003 & 0.963 & 0.010 & 0.975 & 0.002 & 0.979 & 0.004 & 0.962 & 0.007 & 0.969 & 0.005 & 0.941 & 0.016 \\
de-hu-pl & 0.966 & 0.009 & \bfseries 0.982 & 0.007 & 0.967 & 0.011 & 0.965 & 0.012 & 0.980 & 0.005 & 0.966 & 0.010 & 0.969 & 0.011 & \bfseries 0.953 & 0.014 \\
de-pl & 0.966 & 0.008 & 0.981 & 0.006 & 0.969 & 0.009 & 0.965 & 0.012 & 0.974 & 0.011 & 0.967 & 0.006 & 0.967 & 0.011 & 0.951 & 0.010 \\
cs-de & 0.966 & 0.004 & 0.980 & 0.002 & \bfseries 0.972 & 0.007 & 0.963 & 0.012 & 0.973 & 0.008 & 0.951 & 0.007 & \bfseries 0.973 & 0.006 & 0.951 & 0.004 \\
cs-de-pl & 0.966 & 0.006 & 0.981 & 0.005 & 0.966 & 0.009 & 0.961 & 0.014 & 0.975 & 0.009 & 0.962 & 0.007 & 0.971 & 0.008 & 0.948 & 0.009 \\
cs-de-hr-pl & 0.965 & 0.008 & 0.981 & 0.004 & 0.965 & 0.010 & 0.974 & 0.005 & 0.973 & 0.007 & 0.962 & 0.008 & 0.966 & 0.010 & 0.945 & 0.017 \\
cs-hr-pl & 0.962 & 0.007 & 0.982 & 0.004 & 0.948 & 0.016 & 0.975 & 0.003 & 0.977 & 0.004 & 0.966 & 0.005 & 0.965 & 0.011 & 0.935 & 0.015 \\
cs-de-hu-pl & 0.962 & 0.008 & 0.980 & 0.008 & 0.964 & 0.011 & 0.957 & 0.021 & 0.973 & 0.005 & 0.959 & 0.009 & 0.966 & 0.013 & 0.945 & 0.014 \\
de-hr-pl & 0.962 & 0.004 & 0.980 & 0.005 & 0.963 & 0.010 & 0.973 & 0.008 & 0.974 & 0.006 & 0.959 & 0.009 & 0.972 & 0.004 & 0.932 & 0.012 \\
cs-de-hr-hu & 0.961 & 0.006 & 0.978 & 0.003 & 0.964 & 0.012 & 0.975 & 0.005 & 0.978 & 0.003 & 0.950 & 0.008 & 0.961 & 0.007 & 0.929 & 0.016 \\
cs-hu-pl & 0.961 & 0.009 & 0.981 & 0.006 & 0.946 & 0.013 & 0.968 & 0.009 & 0.981 & 0.004 & 0.964 & 0.008 & 0.963 & 0.016 & 0.941 & 0.012 \\
de-hr-hu-pl & 0.961 & 0.009 & 0.977 & 0.005 & 0.962 & 0.014 & 0.971 & 0.005 & 0.974 & 0.009 & 0.959 & 0.009 & 0.965 & 0.007 & 0.929 & 0.023 \\
cs-de-hr & 0.961 & 0.004 & 0.980 & 0.005 & 0.965 & 0.008 & 0.972 & 0.004 & 0.976 & 0.003 & 0.948 & 0.011 & 0.967 & 0.008 & 0.923 & 0.006 \\
cs-hr-hu-pl & 0.960 & 0.005 & 0.981 & 0.004 & 0.944 & 0.018 & 0.971 & 0.003 & 0.980 & 0.003 & 0.961 & 0.007 & 0.957 & 0.006 & 0.927 & 0.016 \\
de-hr-hu & 0.959 & 0.007 & 0.975 & 0.008 & 0.964 & 0.011 & 0.973 & 0.005 & 0.978 & 0.002 & 0.951 & 0.005 & 0.957 & 0.009 & 0.922 & 0.018 \\
de-hu & 0.959 & 0.014 & 0.979 & 0.004 & 0.969 & 0.008 & 0.953 & 0.025 & 0.981 & 0.006 & 0.949 & 0.011 & 0.959 & 0.013 & 0.947 & 0.014 \\
cs-de-hu & 0.959 & 0.008 & 0.975 & 0.008 & 0.968 & 0.010 & 0.958 & 0.005 & 0.976 & 0.005 & 0.944 & 0.008 & 0.960 & 0.012 & 0.938 & 0.016 \\
hu-pl & 0.958 & 0.008 & 0.980 & 0.003 & 0.943 & 0.009 & 0.963 & 0.010 & \bfseries 0.982 & 0.005 & 0.963 & 0.003 & 0.959 & 0.012 & 0.940 & 0.014 \\
de-hr & 0.957 & 0.009 & 0.977 & 0.007 & 0.967 & 0.008 & \bfseries 0.977 & 0.005 & 0.973 & 0.004 & 0.950 & 0.008 & 0.965 & 0.009 & 0.910 & 0.025 \\
hr-hu-pl & 0.956 & 0.007 & 0.977 & 0.005 & 0.947 & 0.012 & 0.969 & 0.006 & 0.979 & 0.005 & 0.964 & 0.006 & 0.952 & 0.009 & 0.912 & 0.028 \\
cs-hr-hu & 0.956 & 0.016 & 0.977 & 0.008 & 0.942 & 0.031 & 0.970 & 0.008 & 0.978 & 0.008 & 0.946 & 0.024 & 0.955 & 0.017 & 0.924 & 0.029 \\
hr-pl & 0.955 & 0.011 & 0.978 & 0.005 & 0.941 & 0.022 & 0.976 & 0.004 & 0.974 & 0.006 & 0.966 & 0.006 & 0.957 & 0.012 & 0.919 & 0.034 \\
cs-pl & 0.955 & 0.017 & 0.977 & 0.012 & 0.941 & 0.024 & 0.961 & 0.014 & 0.973 & 0.013 & 0.960 & 0.017 & 0.960 & 0.022 & 0.935 & 0.027 \\
pl & 0.954 & 0.021 & 0.977 & 0.015 & 0.925 & 0.051 & 0.965 & 0.014 & 0.970 & 0.016 & \bfseries 0.968 & 0.003 & 0.958 & 0.028 & 0.945 & 0.017 \\
cs & 0.954 & 0.015 & 0.976 & 0.002 & 0.941 & 0.022 & 0.958 & 0.018 & 0.971 & 0.010 & 0.946 & 0.015 & 0.960 & 0.016 & 0.941 & 0.015 \\
cs-hu & 0.953 & 0.010 & 0.976 & 0.004 & 0.937 & 0.032 & 0.961 & 0.012 & 0.978 & 0.005 & 0.942 & 0.007 & 0.957 & 0.009 & 0.922 & 0.014 \\
cs-hr & 0.948 & 0.014 & 0.975 & 0.005 & 0.937 & 0.034 & 0.967 & 0.004 & 0.974 & 0.008 & 0.944 & 0.014 & 0.954 & 0.020 & 0.902 & 0.032 \\
hr-hu & 0.947 & 0.020 & 0.972 & 0.010 & 0.937 & 0.030 & 0.970 & 0.007 & 0.981 & 0.005 & 0.941 & 0.017 & 0.945 & 0.026 & 0.905 & 0.042 \\
de & 0.946 & 0.022 & 0.968 & 0.011 & 0.970 & 0.007 & 0.933 & 0.033 & 0.961 & 0.019 & 0.932 & 0.020 & 0.959 & 0.014 & 0.934 & 0.024 \\
hu & 0.934 & 0.030 & 0.968 & 0.013 & 0.927 & 0.038 & 0.941 & 0.037 & 0.980 & 0.006 & 0.929 & 0.028 & 0.934 & 0.016 & 0.910 & 0.027 \\
hr & 0.933 & 0.023 & 0.971 & 0.010 & 0.923 & 0.037 & 0.971 & 0.007 & 0.965 & 0.020 & 0.936 & 0.017 & 0.942 & 0.023 & 0.859 & 0.058 \\
\hline
\end{tabular}
}
\caption{Per-test-language comparison of performance (AUC ROC averaged across the finetuned base models) of finetuned MGT detectors based on combination of train languages. Bold represents the highest value per each test language.}
\label{tab:trainlang}
\vspace{-3mm}
\end{table*}

\subsection{Training Languages Evaluation}

To evaluate effect of a combination of training languages in the model finetuning on MGT detection, we provide the mean performance (AUC ROC) per each train langauges combination, which is averaged across the four base models used for finetuning. These mean values along with standard deviations per each test language are reported in Table~\ref{tab:trainlang}. The results are sorted based on the mean performance for all test langauges combined (All). 

\textbf{There are differences (some of which are statistically significant) in the MGT performance among the combinations of train languages for finetuning.} The results indicate that German and Polish are very important to include in finetuning (\textit{de} is in 9 of the top 10 performing language combinations and \textit{pl} is in 8 of the top 10). However, it is also important to combine at least two train languages, as all of the single-train-language versions of the finetuned models ranked in the bottom 10. In 8 of top 10 versions, at least three languages are combined. A paired t-test ($\alpha$ = 0.05) identified approximately one third of differences between these train-languages combinations as statistically significant (e.g., the top-performing vs. worst-performing combination among them).

It seems that the generalization to Slovenian (i.e., cross-lingual transfer due to Slovenian not present in the training) is the most difficult, requiring Polish or Czech to be present in the finetuning to reach the best performance. Surprisingly, \textbf{Croatian has clearly the worst cross-lingual transferability to Slovenian}, although being the language of the neighboring country of Slovenia and being from the same language-family branch.

There is a small variability across the base models, where the standard deviation is reaching up to 6\% (in most cases under 1\%). The difference between the top-performing combination of train languages and the worst-performing combination is about 3\%. Most of the differences are not statistically significant. Therefore, all of these indicate the performance is quite stable. It seems that it does not matter as much on the base model selected for finetuning as to include a combination of at least two languages for training (ideally from different language families).

\begin{table*}[!t]
\centering
\resizebox{\textwidth}{!}{
\addtolength{\tabcolsep}{-2pt}
\begin{tabular}{c|l||c|c|c|c|c|c|c|c}
\hline
\textbf{Category} & \textbf{Detector} & \bfseries All & \bfseries cs & \bfseries de & \bfseries hr & \bfseries hu & \bfseries pl & \bfseries sk & \bfseries sl \\
\hline
F & Llama-3.2-3B (de-pl-hu) & \bfseries {\cellcolor[HTML]{7EADD1}} \color[HTML]{000000} 0.9758 & \bfseries {\cellcolor[HTML]{78ABD0}} \color[HTML]{000000} 0.9886 & \bfseries {\cellcolor[HTML]{7EADD1}} \color[HTML]{000000} 0.9739 & \bfseries {\cellcolor[HTML]{7BACD1}} \color[HTML]{000000} 0.9829 & \bfseries {\cellcolor[HTML]{79ABD0}} \color[HTML]{000000} 0.9851 & \bfseries {\cellcolor[HTML]{7EADD1}} \color[HTML]{000000} 0.9765 & \bfseries {\cellcolor[HTML]{7DACD1}} \color[HTML]{000000} 0.9779 & \bfseries {\cellcolor[HTML]{83AFD3}} \color[HTML]{000000} 0.9638 \\
F & mDeBERTa-v3-base (de-pl-hr-cs) & {\cellcolor[HTML]{7EADD1}} \color[HTML]{000000} 0.9739 & {\cellcolor[HTML]{7BACD1}} \color[HTML]{000000} 0.9835 & {\cellcolor[HTML]{7EADD1}} \color[HTML]{000000} 0.9731 & {\cellcolor[HTML]{7DACD1}} \color[HTML]{000000} 0.9789 & {\cellcolor[HTML]{7BACD1}} \color[HTML]{000000} 0.9821 & {\cellcolor[HTML]{7EADD1}} \color[HTML]{000000} 0.9727 & {\cellcolor[HTML]{80AED2}} \color[HTML]{000000} 0.9693 & {\cellcolor[HTML]{83AFD3}} \color[HTML]{000000} 0.9624 \\
F & Gemma-2-2B (de-pl-cs) & {\cellcolor[HTML]{81AED2}} \color[HTML]{000000} 0.9660 & {\cellcolor[HTML]{7BACD1}} \color[HTML]{000000} 0.9837 & {\cellcolor[HTML]{80AED2}} \color[HTML]{000000} 0.9697 & {\cellcolor[HTML]{86B0D3}} \color[HTML]{000000} 0.9546 & {\cellcolor[HTML]{7EADD1}} \color[HTML]{000000} 0.9750 & {\cellcolor[HTML]{83AFD3}} \color[HTML]{000000} 0.9620 & {\cellcolor[HTML]{7EADD1}} \color[HTML]{000000} 0.9765 & {\cellcolor[HTML]{8BB2D4}} \color[HTML]{000000} 0.9434 \\
F & XLM-RoBERTa-base (de-pl-hr-hu-cs) & {\cellcolor[HTML]{83AFD3}} \color[HTML]{000000} 0.9621 & {\cellcolor[HTML]{7DACD1}} \color[HTML]{000000} 0.9778 & {\cellcolor[HTML]{89B1D4}} \color[HTML]{000000} 0.9484 & {\cellcolor[HTML]{7EADD1}} \color[HTML]{000000} 0.9744 & {\cellcolor[HTML]{7EADD1}} \color[HTML]{000000} 0.9748 & {\cellcolor[HTML]{86B0D3}} \color[HTML]{000000} 0.9541 & {\cellcolor[HTML]{84B0D3}} \color[HTML]{000000} 0.9606 & {\cellcolor[HTML]{88B1D4}} \color[HTML]{000000} 0.9497 \\
S & Fast-DetectGPT & {\cellcolor[HTML]{C2CBE2}} \color[HTML]{000000} 0.7904 & {\cellcolor[HTML]{C1CAE2}} \color[HTML]{000000} 0.7943 & {\cellcolor[HTML]{BDC8E1}} \color[HTML]{000000} 0.8074 & {\cellcolor[HTML]{B8C6E0}} \color[HTML]{000000} 0.8228 & {\cellcolor[HTML]{CDD0E5}} \color[HTML]{000000} 0.7587 & {\cellcolor[HTML]{C5CCE3}} \color[HTML]{000000} 0.7830 & {\cellcolor[HTML]{C5CCE3}} \color[HTML]{000000} 0.7829 & {\cellcolor[HTML]{BBC7E0}} \color[HTML]{000000} 0.8133 \\
S & Binoculars & {\cellcolor[HTML]{CACEE5}} \color[HTML]{000000} 0.7675 & {\cellcolor[HTML]{C6CCE3}} \color[HTML]{000000} 0.7811 & {\cellcolor[HTML]{C2CBE2}} \color[HTML]{000000} 0.7924 & {\cellcolor[HTML]{C0C9E2}} \color[HTML]{000000} 0.8000 & {\cellcolor[HTML]{D0D1E6}} \color[HTML]{000000} 0.7517 & {\cellcolor[HTML]{CACEE5}} \color[HTML]{000000} 0.7681 & {\cellcolor[HTML]{D1D2E6}} \color[HTML]{000000} 0.7496 & {\cellcolor[HTML]{CCCFE5}} \color[HTML]{000000} 0.7650 \\
S & LLM-Deviation & {\cellcolor[HTML]{DEDCEC}} \color[HTML]{000000} 0.6887 & {\cellcolor[HTML]{CED0E6}} \color[HTML]{000000} 0.7543 & {\cellcolor[HTML]{E3E0EE}} \color[HTML]{000000} 0.6666 & {\cellcolor[HTML]{D6D6E9}} \color[HTML]{000000} 0.7258 & {\cellcolor[HTML]{DFDDEC}} \color[HTML]{000000} 0.6855 & {\cellcolor[HTML]{DCDAEB}} \color[HTML]{000000} 0.6991 & {\cellcolor[HTML]{D9D8EA}} \color[HTML]{000000} 0.7141 & {\cellcolor[HTML]{D4D4E8}} \color[HTML]{000000} 0.7337 \\
P & BLOOMZ-3B-mixed-detector & {\cellcolor[HTML]{E1DFED}} \color[HTML]{000000} 0.6740 & {\cellcolor[HTML]{E0DEED}} \color[HTML]{000000} 0.6769 & {\cellcolor[HTML]{DDDBEC}} \color[HTML]{000000} 0.6945 & {\cellcolor[HTML]{E2DFEE}} \color[HTML]{000000} 0.6690 & {\cellcolor[HTML]{E0DDED}} \color[HTML]{000000} 0.6836 & {\cellcolor[HTML]{E1DFED}} \color[HTML]{000000} 0.6752 & {\cellcolor[HTML]{D5D5E8}} \color[HTML]{000000} 0.7292 & {\cellcolor[HTML]{F0EAF4}} \color[HTML]{000000} 0.5997 \\
P & ChatGPT-detector-RoBERTa-Chinese & {\cellcolor[HTML]{E7E3F0}} \color[HTML]{000000} 0.6492 & {\cellcolor[HTML]{F0EAF4}} \color[HTML]{000000} 0.6045 & {\cellcolor[HTML]{DAD9EA}} \color[HTML]{000000} 0.7055 & {\cellcolor[HTML]{E8E4F0}} \color[HTML]{000000} 0.6442 & {\cellcolor[HTML]{D6D6E9}} \color[HTML]{000000} 0.7238 & {\cellcolor[HTML]{EAE6F1}} \color[HTML]{000000} 0.6361 & {\cellcolor[HTML]{DEDCEC}} \color[HTML]{000000} 0.6885 & {\cellcolor[HTML]{E4E1EF}} \color[HTML]{000000} 0.6629 \\
P & Detection-Longformer & {\cellcolor[HTML]{F5EFF6}} \color[HTML]{000000} 0.5629 & {\cellcolor[HTML]{F5EEF6}} \color[HTML]{000000} 0.5687 & {\cellcolor[HTML]{FFF7FB}} \color[HTML]{000000} 0.4860 & {\cellcolor[HTML]{E7E3F0}} \color[HTML]{000000} 0.6519 & {\cellcolor[HTML]{EBE6F2}} \color[HTML]{000000} 0.6319 & {\cellcolor[HTML]{F1EBF5}} \color[HTML]{000000} 0.5900 & {\cellcolor[HTML]{FFF7FB}} \color[HTML]{000000} 0.4636 & {\cellcolor[HTML]{F5EFF6}} \color[HTML]{000000} 0.5654 \\
\hline
\end{tabular}
}
\caption{Per-test-language comparison of performance (AUC ROC) of categories of MGT detectors (S -- statistical, P -- pretrained, F -- finetuned). For readability, the finetuned category includes only the best performing combination of train languages of each base model. Bold represents the highest value per each test language.}
\label{tab:categories}
\end{table*}

\begin{table*}[!t]
\centering
\resizebox{\textwidth}{!}{
\addtolength{\tabcolsep}{-2pt}
\begin{tabular}{c|l||c|c|c|c|c|c|c|c|c}
\hline
\bfseries Category & \bfseries Detector & \bfseries All & \bfseries \rotatebox{90}{\parbox[c]{2.5cm}{Llama-2-70B-Chat-HF}} & \bfseries \rotatebox{90}{\parbox[c]{2.5cm}{Mistral-7B-Instruct-v0.2}} & \bfseries \rotatebox{90}{Aya-101} & \bfseries \rotatebox{90}{Gemini} & \bfseries \rotatebox{90}{\parbox[c]{2.5cm}{GPT-3.5-Turbo-0125}} & \bfseries \rotatebox{90}{\parbox[c]{2.5cm}{OPT-IML-Max-30B}} & \bfseries \rotatebox{90}{v5-Eagle-7B-HF} & \bfseries \rotatebox{90}{Vicuna-13B} \\
\hline
F & Llama-3.2-3B (de-pl-hu) & \bfseries {\cellcolor[HTML]{7EADD1}} \color[HTML]{000000} 0.9749 & {\cellcolor[HTML]{78ABD0}} \color[HTML]{000000} 0.9904 & {\cellcolor[HTML]{80AED2}} \color[HTML]{000000} 0.9688 & {\cellcolor[HTML]{81AED2}} \color[HTML]{000000} 0.9655 & \bfseries {\cellcolor[HTML]{7EADD1}} \color[HTML]{000000} 0.9754 & \bfseries {\cellcolor[HTML]{7BACD1}} \color[HTML]{000000} 0.9828 & \bfseries {\cellcolor[HTML]{83AFD3}} \color[HTML]{000000} 0.9632 & {\cellcolor[HTML]{7BACD1}} \color[HTML]{000000} 0.9822 & {\cellcolor[HTML]{7DACD1}} \color[HTML]{000000} 0.9789 \\
F & mDeBERTa-v3-base (de-pl-hr-hu-cs) & {\cellcolor[HTML]{7EADD1}} \color[HTML]{000000} 0.9734 & \bfseries {\cellcolor[HTML]{76AAD0}} \color[HTML]{000000} 0.9935 & \bfseries {\cellcolor[HTML]{80AED2}} \color[HTML]{000000} 0.9704 & \bfseries {\cellcolor[HTML]{81AED2}} \color[HTML]{000000} 0.9662 & {\cellcolor[HTML]{80AED2}} \color[HTML]{000000} 0.9706 & {\cellcolor[HTML]{7DACD1}} \color[HTML]{000000} 0.9797 & {\cellcolor[HTML]{88B1D4}} \color[HTML]{000000} 0.9505 & \bfseries {\cellcolor[HTML]{79ABD0}} \color[HTML]{000000} 0.9848 & \bfseries {\cellcolor[HTML]{7DACD1}} \color[HTML]{000000} 0.9801 \\
F & Gemma-2-2B (de-pl-cs) & {\cellcolor[HTML]{83AFD3}} \color[HTML]{000000} 0.9627 & {\cellcolor[HTML]{7DACD1}} \color[HTML]{000000} 0.9777 & {\cellcolor[HTML]{86B0D3}} \color[HTML]{000000} 0.9555 & {\cellcolor[HTML]{83AFD3}} \color[HTML]{000000} 0.9612 & {\cellcolor[HTML]{84B0D3}} \color[HTML]{000000} 0.9585 & {\cellcolor[HTML]{80AED2}} \color[HTML]{000000} 0.9717 & {\cellcolor[HTML]{8BB2D4}} \color[HTML]{000000} 0.9452 & {\cellcolor[HTML]{80AED2}} \color[HTML]{000000} 0.9722 & {\cellcolor[HTML]{81AED2}} \color[HTML]{000000} 0.9653 \\
F & XLM-RoBERTa-base (de-pl-hr-hu-cs) & {\cellcolor[HTML]{83AFD3}} \color[HTML]{000000} 0.9613 & {\cellcolor[HTML]{7BACD1}} \color[HTML]{000000} 0.9821 & {\cellcolor[HTML]{8BB2D4}} \color[HTML]{000000} 0.9430 & {\cellcolor[HTML]{88B1D4}} \color[HTML]{000000} 0.9526 & {\cellcolor[HTML]{83AFD3}} \color[HTML]{000000} 0.9626 & {\cellcolor[HTML]{7EADD1}} \color[HTML]{000000} 0.9745 & {\cellcolor[HTML]{8CB3D5}} \color[HTML]{000000} 0.9403 & {\cellcolor[HTML]{7EADD1}} \color[HTML]{000000} 0.9732 & {\cellcolor[HTML]{7EADD1}} \color[HTML]{000000} 0.9732 \\
S & Fast-DetectGPT & {\cellcolor[HTML]{C5CCE3}} \color[HTML]{000000} 0.7830 & {\cellcolor[HTML]{81AED2}} \color[HTML]{000000} 0.9667 & {\cellcolor[HTML]{EBE6F2}} \color[HTML]{000000} 0.6305 & {\cellcolor[HTML]{B5C4DF}} \color[HTML]{000000} 0.8292 & {\cellcolor[HTML]{C2CBE2}} \color[HTML]{000000} 0.7920 & {\cellcolor[HTML]{C0C9E2}} \color[HTML]{000000} 0.7986 & {\cellcolor[HTML]{E9E5F1}} \color[HTML]{000000} 0.6377 & {\cellcolor[HTML]{A7BDDB}} \color[HTML]{000000} 0.8729 & {\cellcolor[HTML]{B4C4DF}} \color[HTML]{000000} 0.8325 \\
S & Binoculars & {\cellcolor[HTML]{CDD0E5}} \color[HTML]{000000} 0.7603 & {\cellcolor[HTML]{86B0D3}} \color[HTML]{000000} 0.9555 & {\cellcolor[HTML]{EDE7F2}} \color[HTML]{000000} 0.6211 & {\cellcolor[HTML]{C1CAE2}} \color[HTML]{000000} 0.7955 & {\cellcolor[HTML]{C4CBE3}} \color[HTML]{000000} 0.7889 & {\cellcolor[HTML]{C8CDE4}} \color[HTML]{000000} 0.7756 & {\cellcolor[HTML]{F0EAF4}} \color[HTML]{000000} 0.5985 & {\cellcolor[HTML]{ADC1DD}} \color[HTML]{000000} 0.8518 & {\cellcolor[HTML]{BDC8E1}} \color[HTML]{000000} 0.8075 \\
S & LLM-Deviation & {\cellcolor[HTML]{DFDDEC}} \color[HTML]{000000} 0.6843 & {\cellcolor[HTML]{A1BBDA}} \color[HTML]{000000} 0.8873 & {\cellcolor[HTML]{EDE7F2}} \color[HTML]{000000} 0.6233 & {\cellcolor[HTML]{E0DDED}} \color[HTML]{000000} 0.6807 & {\cellcolor[HTML]{E5E1EF}} \color[HTML]{000000} 0.6584 & {\cellcolor[HTML]{DFDDEC}} \color[HTML]{000000} 0.6861 & {\cellcolor[HTML]{F6EFF7}} \color[HTML]{000000} 0.5610 & {\cellcolor[HTML]{CED0E6}} \color[HTML]{000000} 0.7570 & {\cellcolor[HTML]{D9D8EA}} \color[HTML]{000000} 0.7088 \\
P & BLOOMZ-3B-mixed-detector & {\cellcolor[HTML]{E1DFED}} \color[HTML]{000000} 0.6730 & {\cellcolor[HTML]{F9F2F8}} \color[HTML]{000000} 0.5410 & {\cellcolor[HTML]{EBE6F2}} \color[HTML]{000000} 0.6302 & {\cellcolor[HTML]{E3E0EE}} \color[HTML]{000000} 0.6655 & {\cellcolor[HTML]{ECE7F2}} \color[HTML]{000000} 0.6271 & {\cellcolor[HTML]{CACEE5}} \color[HTML]{000000} 0.7694 & {\cellcolor[HTML]{E2DFEE}} \color[HTML]{000000} 0.6697 & {\cellcolor[HTML]{DBDAEB}} \color[HTML]{000000} 0.7000 & {\cellcolor[HTML]{DDDBEC}} \color[HTML]{000000} 0.6923 \\
P & ChatGPT-detector-RoBERTa-Chinese & {\cellcolor[HTML]{E8E4F0}} \color[HTML]{000000} 0.6423 & {\cellcolor[HTML]{C1CAE2}} \color[HTML]{000000} 0.7949 & {\cellcolor[HTML]{E0DEED}} \color[HTML]{000000} 0.6784 & {\cellcolor[HTML]{EFE9F3}} \color[HTML]{000000} 0.6090 & {\cellcolor[HTML]{DEDCEC}} \color[HTML]{000000} 0.6906 & {\cellcolor[HTML]{F2ECF5}} \color[HTML]{000000} 0.5882 & {\cellcolor[HTML]{F5EEF6}} \color[HTML]{000000} 0.5667 & {\cellcolor[HTML]{E9E5F1}} \color[HTML]{000000} 0.6403 & {\cellcolor[HTML]{E2DFEE}} \color[HTML]{000000} 0.6711 \\
P & Detection-Longformer & {\cellcolor[HTML]{F6EFF7}} \color[HTML]{000000} 0.5615 & {\cellcolor[HTML]{E5E1EF}} \color[HTML]{000000} 0.6601 & {\cellcolor[HTML]{F0EAF4}} \color[HTML]{000000} 0.6045 & {\cellcolor[HTML]{FCF4FA}} \color[HTML]{000000} 0.5209 & {\cellcolor[HTML]{FEF6FB}} \color[HTML]{000000} 0.5045 & {\cellcolor[HTML]{FFF7FB}} \color[HTML]{000000} 0.4613 & {\cellcolor[HTML]{F8F1F8}} \color[HTML]{000000} 0.5453 & {\cellcolor[HTML]{ECE7F2}} \color[HTML]{000000} 0.6283 & {\cellcolor[HTML]{F2ECF5}} \color[HTML]{000000} 0.5879 \\
\hline
\end{tabular}
}
\caption{Per-generator comparison of performance (AUC ROC) of categories of MGT detectors (S -- statistical, P -- pretrained, F -- finetuned). For readability, the finetuned category includes only the best performing combination of train languages of each base model. Bold represents the highest value per each MGT generator.}
\label{tab:categories_generators}
\end{table*}

\subsection{Detectors Categories Comparison}

We have compared the performance of three different categories of MGT detectors, namely statistical, pretrained, and finetuned. The comparison of the detectors is provided in Table~\ref{tab:categories}. In the table, we are showing only the best performing combination of train languages for each base model for better readability and space limitation (the worst performing combination is not significantly lower as can be seen in Table~\ref{tab:trainlang}). The comparison of the selected MGT detectors shows that the \textbf{finetuned detectors are consistently the best performing category across all languages}, followed by statistical methods. The worst performing are pretrained detectors, for which this Central European set of languages might be too out-of-distribution. The finetuned detectors achieve by more than 10\% higher performance than the statistical detectors. The difference between statistical and pretrained categories is not that high, for some languages some pretrained detectors outperformed the worst of statistical detectors (LLM-Deviation).

When looking at the results per each MGT generation model, provided in Table~\ref{tab:categories_generators}, we can observe differences especially for statistical detectors. They are best at detecting Llama-2 generated texts (around 0.9 of AUC ROC); on the other hand, Mistral or OPT-IML data are the hardest for them (around 0.6 of AUC ROC). The further analysis is needed to explore this phenomenon. We can speculate that this is due to worsen data quality of MGTs generated by these two models in Central European languages, but there can be a relationship between the mGPT model used as a base model for statistical detectors and the generators (either with the good performance or those two with the low performance).

\begin{table*}[!t]
\centering
\resizebox{\textwidth}{!}{
\addtolength{\tabcolsep}{-2pt}
\begin{tabular}{c|l||c|c|c|c|c|c|c|c}
\hline
\textbf{Domain} & \textbf{Detector} & \bfseries All & \bfseries cs & \bfseries de & \bfseries hr & \bfseries hu & \bfseries pl & \bfseries sk & \bfseries sl \\
\hline
\multirow[c]{10}{*}{\rotatebox{90}{\textbf{News}}} & Llama-3.2-3B (hr-hu-cs) & \bfseries \bfseries {\cellcolor[HTML]{76AAD0}} \color[HTML]{000000} 0.9952 & {\cellcolor[HTML]{75A9CF}} \color[HTML]{000000} 0.9976 & \bfseries \bfseries {\cellcolor[HTML]{76AAD0}} \color[HTML]{000000} 0.9926 & \bfseries \bfseries {\cellcolor[HTML]{75A9CF}} \color[HTML]{000000} 0.9994 & \bfseries \bfseries {\cellcolor[HTML]{75A9CF}} \color[HTML]{000000} 0.9967 & \bfseries \bfseries {\cellcolor[HTML]{76AAD0}} \color[HTML]{000000} 0.9937 & {\cellcolor[HTML]{76AAD0}} \color[HTML]{000000} 0.9928 & {\cellcolor[HTML]{76AAD0}} \color[HTML]{000000} 0.9943 \\
 & mDeBERTa-v3-base (cs) & {\cellcolor[HTML]{76AAD0}} \color[HTML]{000000} 0.9940 & \bfseries \bfseries {\cellcolor[HTML]{75A9CF}} \color[HTML]{000000} 0.9986 & {\cellcolor[HTML]{78ABD0}} \color[HTML]{000000} 0.9921 & {\cellcolor[HTML]{76AAD0}} \color[HTML]{000000} 0.9924 & {\cellcolor[HTML]{76AAD0}} \color[HTML]{000000} 0.9925 & {\cellcolor[HTML]{78ABD0}} \color[HTML]{000000} 0.9900 & \bfseries \bfseries {\cellcolor[HTML]{75A9CF}} \color[HTML]{000000} 0.9981 & \bfseries \bfseries {\cellcolor[HTML]{75A9CF}} \color[HTML]{000000} 0.9973 \\
 & Gemma-2-2B (de-pl-cs) & {\cellcolor[HTML]{78ABD0}} \color[HTML]{000000} 0.9911 & {\cellcolor[HTML]{75A9CF}} \color[HTML]{000000} 0.9966 & {\cellcolor[HTML]{78ABD0}} \color[HTML]{000000} 0.9912 & {\cellcolor[HTML]{7BACD1}} \color[HTML]{000000} 0.9827 & {\cellcolor[HTML]{79ABD0}} \color[HTML]{000000} 0.9882 & {\cellcolor[HTML]{79ABD0}} \color[HTML]{000000} 0.9878 & {\cellcolor[HTML]{75A9CF}} \color[HTML]{000000} 0.9978 & {\cellcolor[HTML]{78ABD0}} \color[HTML]{000000} 0.9894 \\
 & XLM-RoBERTa-base (de-pl-hr-hu-cs) & {\cellcolor[HTML]{7BACD1}} \color[HTML]{000000} 0.9824 & {\cellcolor[HTML]{78ABD0}} \color[HTML]{000000} 0.9896 & {\cellcolor[HTML]{7EADD1}} \color[HTML]{000000} 0.9728 & {\cellcolor[HTML]{79ABD0}} \color[HTML]{000000} 0.9876 & {\cellcolor[HTML]{7EADD1}} \color[HTML]{000000} 0.9759 & {\cellcolor[HTML]{79ABD0}} \color[HTML]{000000} 0.9847 & {\cellcolor[HTML]{7DACD1}} \color[HTML]{000000} 0.9769 & {\cellcolor[HTML]{78ABD0}} \color[HTML]{000000} 0.9910 \\
 & Fast-DetectGPT & {\cellcolor[HTML]{AFC1DD}} \color[HTML]{000000} 0.8490 & {\cellcolor[HTML]{A5BDDB}} \color[HTML]{000000} 0.8773 & {\cellcolor[HTML]{A7BDDB}} \color[HTML]{000000} 0.8717 & {\cellcolor[HTML]{A5BDDB}} \color[HTML]{000000} 0.8777 & {\cellcolor[HTML]{C4CBE3}} \color[HTML]{000000} 0.7867 & {\cellcolor[HTML]{B4C4DF}} \color[HTML]{000000} 0.8351 & {\cellcolor[HTML]{B1C2DE}} \color[HTML]{000000} 0.8413 & {\cellcolor[HTML]{99B8D8}} \color[HTML]{000000} 0.9090 \\
 & Binoculars & {\cellcolor[HTML]{B4C4DF}} \color[HTML]{000000} 0.8341 & {\cellcolor[HTML]{A5BDDB}} \color[HTML]{000000} 0.8771 & {\cellcolor[HTML]{ADC1DD}} \color[HTML]{000000} 0.8536 & {\cellcolor[HTML]{A8BEDC}} \color[HTML]{000000} 0.8707 & {\cellcolor[HTML]{C6CCE3}} \color[HTML]{000000} 0.7809 & {\cellcolor[HTML]{B8C6E0}} \color[HTML]{000000} 0.8228 & {\cellcolor[HTML]{B5C4DF}} \color[HTML]{000000} 0.8298 & {\cellcolor[HTML]{9EBAD9}} \color[HTML]{000000} 0.8965 \\
 & LLM-Deviation & {\cellcolor[HTML]{DAD9EA}} \color[HTML]{000000} 0.7060 & {\cellcolor[HTML]{99B8D8}} \color[HTML]{000000} 0.9083 & {\cellcolor[HTML]{D5D5E8}} \color[HTML]{000000} 0.7298 & {\cellcolor[HTML]{9AB8D8}} \color[HTML]{000000} 0.9025 & {\cellcolor[HTML]{E8E4F0}} \color[HTML]{000000} 0.6429 & {\cellcolor[HTML]{D0D1E6}} \color[HTML]{000000} 0.7507 & {\cellcolor[HTML]{BDC8E1}} \color[HTML]{000000} 0.8048 & {\cellcolor[HTML]{99B8D8}} \color[HTML]{000000} 0.9072 \\
 & Detection-Longformer & {\cellcolor[HTML]{E7E3F0}} \color[HTML]{000000} 0.6503 & {\cellcolor[HTML]{EAE6F1}} \color[HTML]{000000} 0.6356 & {\cellcolor[HTML]{EFE9F3}} \color[HTML]{000000} 0.6074 & {\cellcolor[HTML]{CDD0E5}} \color[HTML]{000000} 0.7595 & {\cellcolor[HTML]{D0D1E6}} \color[HTML]{000000} 0.7507 & {\cellcolor[HTML]{DBDAEB}} \color[HTML]{000000} 0.7003 & {\cellcolor[HTML]{FFF7FB}} \color[HTML]{000000} 0.4962 & {\cellcolor[HTML]{D8D7E9}} \color[HTML]{000000} 0.7168 \\
 & ChatGPT-detector-RoBERTa-Chinese & {\cellcolor[HTML]{EDE7F2}} \color[HTML]{000000} 0.6223 & {\cellcolor[HTML]{FAF2F8}} \color[HTML]{000000} 0.5364 & {\cellcolor[HTML]{D8D7E9}} \color[HTML]{000000} 0.7168 & {\cellcolor[HTML]{E3E0EE}} \color[HTML]{000000} 0.6672 & {\cellcolor[HTML]{E2DFEE}} \color[HTML]{000000} 0.6717 & {\cellcolor[HTML]{E6E2EF}} \color[HTML]{000000} 0.6541 & {\cellcolor[HTML]{CCCFE5}} \color[HTML]{000000} 0.7646 & {\cellcolor[HTML]{DAD9EA}} \color[HTML]{000000} 0.7038 \\
 & BLOOMZ-3B-mixed-detector & {\cellcolor[HTML]{F5EFF6}} \color[HTML]{000000} 0.5626 & {\cellcolor[HTML]{FEF6FB}} \color[HTML]{000000} 0.5049 & {\cellcolor[HTML]{F1EBF4}} \color[HTML]{000000} 0.5963 & {\cellcolor[HTML]{FBF4F9}} \color[HTML]{000000} 0.5271 & {\cellcolor[HTML]{FFF7FB}} \color[HTML]{000000} 0.4680 & {\cellcolor[HTML]{F5EEF6}} \color[HTML]{000000} 0.5691 & {\cellcolor[HTML]{DCDAEB}} \color[HTML]{000000} 0.6970 & {\cellcolor[HTML]{F7F0F7}} \color[HTML]{000000} 0.5544 \\
\hline
\multirow[c]{10}{*}{\rotatebox{90}{\textbf{Social media}}} & Llama-3.2-3B (de-pl-hu) & \bfseries {\cellcolor[HTML]{88B1D4}} \color[HTML]{000000} 0.9506 & \bfseries {\cellcolor[HTML]{7DACD1}} \color[HTML]{000000} 0.9800 & {\cellcolor[HTML]{8BB2D4}} \color[HTML]{000000} 0.9427 & \bfseries {\cellcolor[HTML]{83AFD3}} \color[HTML]{000000} 0.9628 & \bfseries {\cellcolor[HTML]{7EADD1}} \color[HTML]{000000} 0.9744 & \bfseries {\cellcolor[HTML]{89B1D4}} \color[HTML]{000000} 0.9466 & \bfseries {\cellcolor[HTML]{88B1D4}} \color[HTML]{000000} 0.9527 & {\cellcolor[HTML]{96B6D7}} \color[HTML]{000000} 0.9176 \\
 & mDeBERTa-v3-base (de) & {\cellcolor[HTML]{89B1D4}} \color[HTML]{000000} 0.9476 & {\cellcolor[HTML]{86B0D3}} \color[HTML]{000000} 0.9536 & \bfseries {\cellcolor[HTML]{88B1D4}} \color[HTML]{000000} 0.9515 & {\cellcolor[HTML]{88B1D4}} \color[HTML]{000000} 0.9503 & {\cellcolor[HTML]{80AED2}} \color[HTML]{000000} 0.9708 & {\cellcolor[HTML]{8CB3D5}} \color[HTML]{000000} 0.9413 & {\cellcolor[HTML]{8BB2D4}} \color[HTML]{000000} 0.9439 & \bfseries {\cellcolor[HTML]{8FB4D6}} \color[HTML]{000000} 0.9306 \\
 & XLM-RoBERTa-base (de-pl) & {\cellcolor[HTML]{8CB3D5}} \color[HTML]{000000} 0.9412 & {\cellcolor[HTML]{84B0D3}} \color[HTML]{000000} 0.9590 & {\cellcolor[HTML]{8EB3D5}} \color[HTML]{000000} 0.9344 & {\cellcolor[HTML]{86B0D3}} \color[HTML]{000000} 0.9548 & {\cellcolor[HTML]{81AED2}} \color[HTML]{000000} 0.9670 & {\cellcolor[HTML]{91B5D6}} \color[HTML]{000000} 0.9282 & {\cellcolor[HTML]{8BB2D4}} \color[HTML]{000000} 0.9424 & {\cellcolor[HTML]{9EBAD9}} \color[HTML]{000000} 0.8972 \\
 & Gemma-2-2B (de-pl-hr-hu-cs) & {\cellcolor[HTML]{8FB4D6}} \color[HTML]{000000} 0.9313 & {\cellcolor[HTML]{83AFD3}} \color[HTML]{000000} 0.9631 & {\cellcolor[HTML]{8FB4D6}} \color[HTML]{000000} 0.9334 & {\cellcolor[HTML]{89B1D4}} \color[HTML]{000000} 0.9468 & {\cellcolor[HTML]{81AED2}} \color[HTML]{000000} 0.9686 & {\cellcolor[HTML]{93B5D6}} \color[HTML]{000000} 0.9240 & {\cellcolor[HTML]{8FB4D6}} \color[HTML]{000000} 0.9324 & {\cellcolor[HTML]{AFC1DD}} \color[HTML]{000000} 0.8483 \\
 & LLM-Deviation & {\cellcolor[HTML]{BDC8E1}} \color[HTML]{000000} 0.8049 & {\cellcolor[HTML]{A1BBDA}} \color[HTML]{000000} 0.8877 & {\cellcolor[HTML]{D5D5E8}} \color[HTML]{000000} 0.7279 & {\cellcolor[HTML]{BFC9E1}} \color[HTML]{000000} 0.8030 & {\cellcolor[HTML]{9CB9D9}} \color[HTML]{000000} 0.8990 & {\cellcolor[HTML]{BBC7E0}} \color[HTML]{000000} 0.8128 & {\cellcolor[HTML]{C5CCE3}} \color[HTML]{000000} 0.7818 & {\cellcolor[HTML]{D1D2E6}} \color[HTML]{000000} 0.7484 \\
 & Binoculars & {\cellcolor[HTML]{C9CEE4}} \color[HTML]{000000} 0.7699 & {\cellcolor[HTML]{C2CBE2}} \color[HTML]{000000} 0.7911 & {\cellcolor[HTML]{C2CBE2}} \color[HTML]{000000} 0.7922 & {\cellcolor[HTML]{BCC7E1}} \color[HTML]{000000} 0.8107 & {\cellcolor[HTML]{C4CBE3}} \color[HTML]{000000} 0.7856 & {\cellcolor[HTML]{CDD0E5}} \color[HTML]{000000} 0.7598 & {\cellcolor[HTML]{D2D3E7}} \color[HTML]{000000} 0.7384 & {\cellcolor[HTML]{D8D7E9}} \color[HTML]{000000} 0.7169 \\
 & BLOOMZ-3B-mixed-detector & {\cellcolor[HTML]{CCCFE5}} \color[HTML]{000000} 0.7627 & {\cellcolor[HTML]{C0C9E2}} \color[HTML]{000000} 0.7989 & {\cellcolor[HTML]{C5CCE3}} \color[HTML]{000000} 0.7843 & {\cellcolor[HTML]{C8CDE4}} \color[HTML]{000000} 0.7748 & {\cellcolor[HTML]{B3C3DE}} \color[HTML]{000000} 0.8394 & {\cellcolor[HTML]{CACEE5}} \color[HTML]{000000} 0.7661 & {\cellcolor[HTML]{CCCFE5}} \color[HTML]{000000} 0.7625 & {\cellcolor[HTML]{E8E4F0}} \color[HTML]{000000} 0.6438 \\
 & Fast-DetectGPT & {\cellcolor[HTML]{CCCFE5}} \color[HTML]{000000} 0.7617 & {\cellcolor[HTML]{CCCFE5}} \color[HTML]{000000} 0.7626 & {\cellcolor[HTML]{C5CCE3}} \color[HTML]{000000} 0.7827 & {\cellcolor[HTML]{BFC9E1}} \color[HTML]{000000} 0.8044 & {\cellcolor[HTML]{C6CCE3}} \color[HTML]{000000} 0.7780 & {\cellcolor[HTML]{D1D2E6}} \color[HTML]{000000} 0.7500 & {\cellcolor[HTML]{D1D2E6}} \color[HTML]{000000} 0.7467 & {\cellcolor[HTML]{D6D6E9}} \color[HTML]{000000} 0.7238 \\
 & ChatGPT-detector-RoBERTa-Chinese & {\cellcolor[HTML]{E1DFED}} \color[HTML]{000000} 0.6737 & {\cellcolor[HTML]{E4E1EF}} \color[HTML]{000000} 0.6605 & {\cellcolor[HTML]{C0C9E2}} \color[HTML]{000000} 0.7974 & {\cellcolor[HTML]{EDE7F2}} \color[HTML]{000000} 0.6211 & {\cellcolor[HTML]{C6CCE3}} \color[HTML]{000000} 0.7778 & {\cellcolor[HTML]{EDE8F3}} \color[HTML]{000000} 0.6179 & {\cellcolor[HTML]{EAE6F1}} \color[HTML]{000000} 0.6345 & {\cellcolor[HTML]{E7E3F0}} \color[HTML]{000000} 0.6480 \\
 & Detection-Longformer & {\cellcolor[HTML]{FFF7FB}} \color[HTML]{000000} 0.4757 & {\cellcolor[HTML]{FEF6FB}} \color[HTML]{000000} 0.5054 & {\cellcolor[HTML]{FFF7FB}} \color[HTML]{000000} 0.3848 & {\cellcolor[HTML]{F8F1F8}} \color[HTML]{000000} 0.5480 & {\cellcolor[HTML]{FBF3F9}} \color[HTML]{000000} 0.5288 & {\cellcolor[HTML]{FFF7FB}} \color[HTML]{000000} 0.4772 & {\cellcolor[HTML]{FFF7FB}} \color[HTML]{000000} 0.4293 & {\cellcolor[HTML]{FFF7FB}} \color[HTML]{000000} 0.4382 \\
\hline
\end{tabular}
}
\caption{Per-test-language comparison of performance (AUC ROC) of the selected MGT detectors for each domain. For readability, the finetuned category includes only the best performing combination of train languages of each base model (for each domain). Bold represents the highest value per each test language and each domain.}
\label{tab:domains}
\end{table*}

\begin{table*}[!t]
\centering
\resizebox{\textwidth}{!}{
\addtolength{\tabcolsep}{-2pt}
\begin{tabular}{c|l||c|c|c|c|c|c|c|c}
\hline
\textbf{Detector} & \textbf{Subset} & \bfseries All & \bfseries cs & \bfseries de & \bfseries hr & \bfseries hu & \bfseries pl & \bfseries sk & \bfseries sl \\
\hline
\multirow[c]{3}{*}{Llama-3.2-3B} & original & {\cellcolor[HTML]{7EADD1}} \color[HTML]{000000} 0.9739 & {\cellcolor[HTML]{7DACD1}} \color[HTML]{000000} 0.9787 & {\cellcolor[HTML]{80AED2}} \color[HTML]{000000} 0.9717 & {\cellcolor[HTML]{79ABD0}} \color[HTML]{000000} 0.9881 & {\cellcolor[HTML]{7BACD1}} \color[HTML]{000000} 0.9825 & {\cellcolor[HTML]{7BACD1}} \color[HTML]{000000} 0.9808 & {\cellcolor[HTML]{80AED2}} \color[HTML]{000000} 0.9703 & {\cellcolor[HTML]{81AED2}} \color[HTML]{000000} 0.9654 \\
 & paraphrased & \bfseries \bfseries {\cellcolor[HTML]{79ABD0}} \color[HTML]{000000} 0.9851 & \bfseries \bfseries {\cellcolor[HTML]{78ABD0}} \color[HTML]{000000} 0.9921 & \bfseries {\cellcolor[HTML]{7DACD1}} \color[HTML]{000000} 0.9785 & \bfseries \bfseries {\cellcolor[HTML]{75A9CF}} \color[HTML]{000000} 0.9966 & \bfseries \bfseries {\cellcolor[HTML]{79ABD0}} \color[HTML]{000000} 0.9869 & \bfseries \bfseries {\cellcolor[HTML]{79ABD0}} \color[HTML]{000000} 0.9867 & \bfseries {\cellcolor[HTML]{79ABD0}} \color[HTML]{000000} 0.9845 & \bfseries \bfseries {\cellcolor[HTML]{7BACD1}} \color[HTML]{000000} 0.9827 \\
 & homoglyph & {\cellcolor[HTML]{8EB3D5}} \color[HTML]{000000} 0.9354 & {\cellcolor[HTML]{83AFD3}} \color[HTML]{000000} 0.9630 & {\cellcolor[HTML]{91B5D6}} \color[HTML]{000000} 0.9282 & {\cellcolor[HTML]{80AED2}} \color[HTML]{000000} 0.9726 & {\cellcolor[HTML]{88B1D4}} \color[HTML]{000000} 0.9507 & {\cellcolor[HTML]{8EB3D5}} \color[HTML]{000000} 0.9347 & {\cellcolor[HTML]{91B5D6}} \color[HTML]{000000} 0.9278 & {\cellcolor[HTML]{96B6D7}} \color[HTML]{000000} 0.9157 \\
\hline
\multirow[c]{3}{*}{mDeBERTa-v3-base} & original & {\cellcolor[HTML]{80AED2}} \color[HTML]{000000} 0.9720 & {\cellcolor[HTML]{7DACD1}} \color[HTML]{000000} 0.9769 & {\cellcolor[HTML]{7EADD1}} \color[HTML]{000000} 0.9734 & {\cellcolor[HTML]{81AED2}} \color[HTML]{000000} 0.9674 & {\cellcolor[HTML]{7BACD1}} \color[HTML]{000000} 0.9816 & {\cellcolor[HTML]{7EADD1}} \color[HTML]{000000} 0.9757 & {\cellcolor[HTML]{80AED2}} \color[HTML]{000000} 0.9721 & {\cellcolor[HTML]{83AFD3}} \color[HTML]{000000} 0.9636 \\
 & paraphrased & \bfseries {\cellcolor[HTML]{7BACD1}} \color[HTML]{000000} 0.9825 & \bfseries {\cellcolor[HTML]{79ABD0}} \color[HTML]{000000} 0.9881 & \bfseries \bfseries {\cellcolor[HTML]{7BACD1}} \color[HTML]{000000} 0.9813 & \bfseries {\cellcolor[HTML]{7DACD1}} \color[HTML]{000000} 0.9782 & \bfseries {\cellcolor[HTML]{79ABD0}} \color[HTML]{000000} 0.9869 & \bfseries {\cellcolor[HTML]{7BACD1}} \color[HTML]{000000} 0.9828 & \bfseries \bfseries {\cellcolor[HTML]{78ABD0}} \color[HTML]{000000} 0.9888 & \bfseries {\cellcolor[HTML]{7EADD1}} \color[HTML]{000000} 0.9744 \\
 & homoglyph & {\cellcolor[HTML]{C8CDE4}} \color[HTML]{000000} 0.7749 & {\cellcolor[HTML]{ADC1DD}} \color[HTML]{000000} 0.8519 & {\cellcolor[HTML]{E1DFED}} \color[HTML]{000000} 0.6736 & {\cellcolor[HTML]{BDC8E1}} \color[HTML]{000000} 0.8080 & {\cellcolor[HTML]{B8C6E0}} \color[HTML]{000000} 0.8215 & {\cellcolor[HTML]{D3D4E7}} \color[HTML]{000000} 0.7367 & {\cellcolor[HTML]{B9C6E0}} \color[HTML]{000000} 0.8170 & {\cellcolor[HTML]{DBDAEB}} \color[HTML]{000000} 0.7024 \\
\hline
\multirow[c]{3}{*}{Gemma-2-2B} & original & {\cellcolor[HTML]{84B0D3}} \color[HTML]{000000} 0.9588 & {\cellcolor[HTML]{7DACD1}} \color[HTML]{000000} 0.9769 & {\cellcolor[HTML]{83AFD3}} \color[HTML]{000000} 0.9647 & {\cellcolor[HTML]{83AFD3}} \color[HTML]{000000} 0.9634 & {\cellcolor[HTML]{7DACD1}} \color[HTML]{000000} 0.9783 & {\cellcolor[HTML]{7EADD1}} \color[HTML]{000000} 0.9752 & {\cellcolor[HTML]{81AED2}} \color[HTML]{000000} 0.9677 & {\cellcolor[HTML]{89B1D4}} \color[HTML]{000000} 0.9484 \\
 & paraphrased & \bfseries {\cellcolor[HTML]{7DACD1}} \color[HTML]{000000} 0.9779 & \bfseries {\cellcolor[HTML]{78ABD0}} \color[HTML]{000000} 0.9889 & \bfseries {\cellcolor[HTML]{7EADD1}} \color[HTML]{000000} 0.9750 & \bfseries {\cellcolor[HTML]{78ABD0}} \color[HTML]{000000} 0.9908 & \bfseries {\cellcolor[HTML]{7BACD1}} \color[HTML]{000000} 0.9809 & \bfseries {\cellcolor[HTML]{7BACD1}} \color[HTML]{000000} 0.9826 & \bfseries {\cellcolor[HTML]{7BACD1}} \color[HTML]{000000} 0.9832 & \bfseries {\cellcolor[HTML]{7EADD1}} \color[HTML]{000000} 0.9741 \\
 & homoglyph & {\cellcolor[HTML]{9EBAD9}} \color[HTML]{000000} 0.8962 & {\cellcolor[HTML]{8CB3D5}} \color[HTML]{000000} 0.9377 & {\cellcolor[HTML]{8FB4D6}} \color[HTML]{000000} 0.9326 & {\cellcolor[HTML]{9AB8D8}} \color[HTML]{000000} 0.9060 & {\cellcolor[HTML]{88B1D4}} \color[HTML]{000000} 0.9530 & {\cellcolor[HTML]{97B7D7}} \color[HTML]{000000} 0.9115 & {\cellcolor[HTML]{9CB9D9}} \color[HTML]{000000} 0.9011 & {\cellcolor[HTML]{ACC0DD}} \color[HTML]{000000} 0.8589 \\
\hline
\multirow[c]{3}{*}{XLM-RoBERTa-base} & original & {\cellcolor[HTML]{86B0D3}} \color[HTML]{000000} 0.9540 & {\cellcolor[HTML]{83AFD3}} \color[HTML]{000000} 0.9633 & {\cellcolor[HTML]{89B1D4}} \color[HTML]{000000} 0.9474 & {\cellcolor[HTML]{83AFD3}} \color[HTML]{000000} 0.9627 & {\cellcolor[HTML]{7EADD1}} \color[HTML]{000000} 0.9736 & {\cellcolor[HTML]{83AFD3}} \color[HTML]{000000} 0.9611 & {\cellcolor[HTML]{8BB2D4}} \color[HTML]{000000} 0.9434 & {\cellcolor[HTML]{8FB4D6}} \color[HTML]{000000} 0.9322 \\
 & paraphrased & \bfseries {\cellcolor[HTML]{81AED2}} \color[HTML]{000000} 0.9665 & \bfseries {\cellcolor[HTML]{7DACD1}} \color[HTML]{000000} 0.9769 & \bfseries {\cellcolor[HTML]{86B0D3}} \color[HTML]{000000} 0.9536 & \bfseries {\cellcolor[HTML]{7DACD1}} \color[HTML]{000000} 0.9774 & \bfseries {\cellcolor[HTML]{7BACD1}} \color[HTML]{000000} 0.9805 & \bfseries {\cellcolor[HTML]{83AFD3}} \color[HTML]{000000} 0.9629 & \bfseries {\cellcolor[HTML]{81AED2}} \color[HTML]{000000} 0.9687 & \bfseries {\cellcolor[HTML]{86B0D3}} \color[HTML]{000000} 0.9536 \\
 & homoglyph & {\cellcolor[HTML]{FCF4FA}} \color[HTML]{000000} 0.5218 & {\cellcolor[HTML]{E9E5F1}} \color[HTML]{000000} 0.6393 & {\cellcolor[HTML]{FFF7FB}} \color[HTML]{000000} 0.3788 & {\cellcolor[HTML]{F7F0F7}} \color[HTML]{000000} 0.5510 & {\cellcolor[HTML]{E9E5F1}} \color[HTML]{000000} 0.6389 & {\cellcolor[HTML]{FBF3F9}} \color[HTML]{000000} 0.5300 & {\cellcolor[HTML]{FFF7FB}} \color[HTML]{000000} 0.4809 & {\cellcolor[HTML]{FFF7FB}} \color[HTML]{000000} 0.4128 \\
\hline
\multirow[c]{3}{*}{Fast-DetectGPT} & original & \bfseries {\cellcolor[HTML]{C0C9E2}} \color[HTML]{000000} 0.8000 & \bfseries {\cellcolor[HTML]{BFC9E1}} \color[HTML]{000000} 0.8045 & \bfseries {\cellcolor[HTML]{BCC7E1}} \color[HTML]{000000} 0.8122 & \bfseries {\cellcolor[HTML]{BDC8E1}} \color[HTML]{000000} 0.8049 & \bfseries {\cellcolor[HTML]{BBC7E0}} \color[HTML]{000000} 0.8154 & \bfseries {\cellcolor[HTML]{BCC7E1}} \color[HTML]{000000} 0.8096 & \bfseries {\cellcolor[HTML]{CACEE5}} \color[HTML]{000000} 0.7658 & \bfseries {\cellcolor[HTML]{BCC7E1}} \color[HTML]{000000} 0.8112 \\
 & paraphrased & {\cellcolor[HTML]{D9D8EA}} \color[HTML]{000000} 0.7074 & {\cellcolor[HTML]{CACEE5}} \color[HTML]{000000} 0.7657 & {\cellcolor[HTML]{C2CBE2}} \color[HTML]{000000} 0.7921 & {\cellcolor[HTML]{D6D6E9}} \color[HTML]{000000} 0.7229 & {\cellcolor[HTML]{FBF3F9}} \color[HTML]{000000} 0.5298 & {\cellcolor[HTML]{DCDAEB}} \color[HTML]{000000} 0.6965 & {\cellcolor[HTML]{D4D4E8}} \color[HTML]{000000} 0.7343 & {\cellcolor[HTML]{D2D3E7}} \color[HTML]{000000} 0.7394 \\
 & homoglyph & {\cellcolor[HTML]{FFF7FB}} \color[HTML]{000000} 0.0815 & {\cellcolor[HTML]{FFF7FB}} \color[HTML]{000000} 0.1093 & {\cellcolor[HTML]{FFF7FB}} \color[HTML]{000000} 0.0403 & {\cellcolor[HTML]{FFF7FB}} \color[HTML]{000000} 0.0542 & {\cellcolor[HTML]{FFF7FB}} \color[HTML]{000000} 0.0808 & {\cellcolor[HTML]{FFF7FB}} \color[HTML]{000000} 0.0857 & {\cellcolor[HTML]{FFF7FB}} \color[HTML]{000000} 0.1232 & {\cellcolor[HTML]{FFF7FB}} \color[HTML]{000000} 0.0750 \\
\hline
\multirow[c]{3}{*}{Binoculars} & original & \bfseries {\cellcolor[HTML]{C8CDE4}} \color[HTML]{000000} 0.7758 & \bfseries {\cellcolor[HTML]{C2CBE2}} \color[HTML]{000000} 0.7894 & \bfseries {\cellcolor[HTML]{C2CBE2}} \color[HTML]{000000} 0.7917 & \bfseries {\cellcolor[HTML]{C8CDE4}} \color[HTML]{000000} 0.7737 & \bfseries {\cellcolor[HTML]{BDC8E1}} \color[HTML]{000000} 0.8080 & \bfseries {\cellcolor[HTML]{C0C9E2}} \color[HTML]{000000} 0.7993 & \bfseries {\cellcolor[HTML]{D3D4E7}} \color[HTML]{000000} 0.7376 & \bfseries {\cellcolor[HTML]{CDD0E5}} \color[HTML]{000000} 0.7601 \\
 & paraphrased & {\cellcolor[HTML]{D9D8EA}} \color[HTML]{000000} 0.7098 & {\cellcolor[HTML]{CACEE5}} \color[HTML]{000000} 0.7683 & {\cellcolor[HTML]{C4CBE3}} \color[HTML]{000000} 0.7855 & {\cellcolor[HTML]{D8D7E9}} \color[HTML]{000000} 0.7171 & {\cellcolor[HTML]{F1EBF4}} \color[HTML]{000000} 0.5951 & {\cellcolor[HTML]{D7D6E9}} \color[HTML]{000000} 0.7188 & {\cellcolor[HTML]{D7D6E9}} \color[HTML]{000000} 0.7219 & {\cellcolor[HTML]{D9D8EA}} \color[HTML]{000000} 0.7121 \\
 & homoglyph & {\cellcolor[HTML]{FFF7FB}} \color[HTML]{000000} 0.2711 & {\cellcolor[HTML]{FFF7FB}} \color[HTML]{000000} 0.2951 & {\cellcolor[HTML]{FFF7FB}} \color[HTML]{000000} 0.2338 & {\cellcolor[HTML]{FFF7FB}} \color[HTML]{000000} 0.2313 & {\cellcolor[HTML]{FFF7FB}} \color[HTML]{000000} 0.3252 & {\cellcolor[HTML]{FFF7FB}} \color[HTML]{000000} 0.2866 & {\cellcolor[HTML]{FFF7FB}} \color[HTML]{000000} 0.2996 & {\cellcolor[HTML]{FFF7FB}} \color[HTML]{000000} 0.2214 \\
\hline
\multirow[c]{3}{*}{LLM-Deviation} & original & \bfseries {\cellcolor[HTML]{DCDAEB}} \color[HTML]{000000} 0.6968 & \bfseries {\cellcolor[HTML]{D1D2E6}} \color[HTML]{000000} 0.7473 & {\cellcolor[HTML]{DCDAEB}} \color[HTML]{000000} 0.6958 & \bfseries {\cellcolor[HTML]{DBDAEB}} \color[HTML]{000000} 0.7030 & \bfseries {\cellcolor[HTML]{D5D5E8}} \color[HTML]{000000} 0.7288 & \bfseries {\cellcolor[HTML]{D7D6E9}} \color[HTML]{000000} 0.7208 & {\cellcolor[HTML]{D9D8EA}} \color[HTML]{000000} 0.7114 & \bfseries {\cellcolor[HTML]{DAD9EA}} \color[HTML]{000000} 0.7035 \\
 & paraphrased & {\cellcolor[HTML]{E0DEED}} \color[HTML]{000000} 0.6775 & {\cellcolor[HTML]{D4D4E8}} \color[HTML]{000000} 0.7322 & \bfseries {\cellcolor[HTML]{DAD9EA}} \color[HTML]{000000} 0.7044 & {\cellcolor[HTML]{DCDAEB}} \color[HTML]{000000} 0.6964 & {\cellcolor[HTML]{F0EAF4}} \color[HTML]{000000} 0.5978 & {\cellcolor[HTML]{E4E1EF}} \color[HTML]{000000} 0.6622 & \bfseries {\cellcolor[HTML]{D8D7E9}} \color[HTML]{000000} 0.7158 & {\cellcolor[HTML]{DEDCEC}} \color[HTML]{000000} 0.6881 \\
 & homoglyph & {\cellcolor[HTML]{FFF7FB}} \color[HTML]{000000} 0.3614 & {\cellcolor[HTML]{FFF7FB}} \color[HTML]{000000} 0.4219 & {\cellcolor[HTML]{FFF7FB}} \color[HTML]{000000} 0.2454 & {\cellcolor[HTML]{FFF7FB}} \color[HTML]{000000} 0.3539 & {\cellcolor[HTML]{FFF7FB}} \color[HTML]{000000} 0.4082 & {\cellcolor[HTML]{FFF7FB}} \color[HTML]{000000} 0.3527 & {\cellcolor[HTML]{FFF7FB}} \color[HTML]{000000} 0.3466 & {\cellcolor[HTML]{FFF7FB}} \color[HTML]{000000} 0.3136 \\
\hline
\multirow[c]{3}{*}{BLOOMz-3B-mixed-detector} & original & {\cellcolor[HTML]{E0DEED}} \color[HTML]{000000} 0.6778 & {\cellcolor[HTML]{E5E1EF}} \color[HTML]{000000} 0.6565 & {\cellcolor[HTML]{E2DFEE}} \color[HTML]{000000} 0.6707 & {\cellcolor[HTML]{DBDAEB}} \color[HTML]{000000} 0.7015 & {\cellcolor[HTML]{DDDBEC}} \color[HTML]{000000} 0.6923 & {\cellcolor[HTML]{DEDCEC}} \color[HTML]{000000} 0.6879 & {\cellcolor[HTML]{D9D8EA}} \color[HTML]{000000} 0.7095 & {\cellcolor[HTML]{E7E3F0}} \color[HTML]{000000} 0.6450 \\
 & paraphrased & \bfseries {\cellcolor[HTML]{AFC1DD}} \color[HTML]{000000} 0.8487 & \bfseries {\cellcolor[HTML]{ADC1DD}} \color[HTML]{000000} 0.8526 & \bfseries {\cellcolor[HTML]{B3C3DE}} \color[HTML]{000000} 0.8379 & \bfseries {\cellcolor[HTML]{ADC1DD}} \color[HTML]{000000} 0.8527 & \bfseries {\cellcolor[HTML]{B4C4DF}} \color[HTML]{000000} 0.8337 & \bfseries {\cellcolor[HTML]{A5BDDB}} \color[HTML]{000000} 0.8754 & \bfseries {\cellcolor[HTML]{A1BBDA}} \color[HTML]{000000} 0.8904 & \bfseries {\cellcolor[HTML]{B7C5DF}} \color[HTML]{000000} 0.8275 \\
 & homoglyph & {\cellcolor[HTML]{FFF7FB}} \color[HTML]{000000} 0.4240 & {\cellcolor[HTML]{FFF7FB}} \color[HTML]{000000} 0.4333 & {\cellcolor[HTML]{FFF7FB}} \color[HTML]{000000} 0.4203 & {\cellcolor[HTML]{FFF7FB}} \color[HTML]{000000} 0.4280 & {\cellcolor[HTML]{FFF7FB}} \color[HTML]{000000} 0.4281 & {\cellcolor[HTML]{FFF7FB}} \color[HTML]{000000} 0.4369 & {\cellcolor[HTML]{FFF7FB}} \color[HTML]{000000} 0.4562 & {\cellcolor[HTML]{FFF7FB}} \color[HTML]{000000} 0.3627 \\
\hline
\multirow[c]{3}{*}{ChatGPT-detector-RoBERTa-Chinese} & original & {\cellcolor[HTML]{E7E3F0}} \color[HTML]{000000} 0.6497 & \bfseries {\cellcolor[HTML]{EDE7F2}} \color[HTML]{000000} 0.6228 & \bfseries {\cellcolor[HTML]{DDDBEC}} \color[HTML]{000000} 0.6915 & {\cellcolor[HTML]{E7E3F0}} \color[HTML]{000000} 0.6473 & {\cellcolor[HTML]{D6D6E9}} \color[HTML]{000000} 0.7228 & \bfseries {\cellcolor[HTML]{EBE6F2}} \color[HTML]{000000} 0.6294 & {\cellcolor[HTML]{E3E0EE}} \color[HTML]{000000} 0.6655 & \bfseries {\cellcolor[HTML]{E0DEED}} \color[HTML]{000000} 0.6769 \\
 & paraphrased & \bfseries {\cellcolor[HTML]{E7E3F0}} \color[HTML]{000000} 0.6515 & {\cellcolor[HTML]{F0EAF4}} \color[HTML]{000000} 0.6008 & {\cellcolor[HTML]{E0DDED}} \color[HTML]{000000} 0.6829 & \bfseries {\cellcolor[HTML]{E1DFED}} \color[HTML]{000000} 0.6741 & \bfseries {\cellcolor[HTML]{D2D3E7}} \color[HTML]{000000} 0.7408 & {\cellcolor[HTML]{EDE8F3}} \color[HTML]{000000} 0.6196 & \bfseries {\cellcolor[HTML]{DCDAEB}} \color[HTML]{000000} 0.6955 & {\cellcolor[HTML]{ECE7F2}} \color[HTML]{000000} 0.6278 \\
 & homoglyph & {\cellcolor[HTML]{F7F0F7}} \color[HTML]{000000} 0.5547 & {\cellcolor[HTML]{FDF5FA}} \color[HTML]{000000} 0.5127 & {\cellcolor[HTML]{F9F2F8}} \color[HTML]{000000} 0.5404 & {\cellcolor[HTML]{FDF5FA}} \color[HTML]{000000} 0.5184 & {\cellcolor[HTML]{EDE7F2}} \color[HTML]{000000} 0.6221 & {\cellcolor[HTML]{F8F1F8}} \color[HTML]{000000} 0.5453 & {\cellcolor[HTML]{F7F0F7}} \color[HTML]{000000} 0.5516 & {\cellcolor[HTML]{FEF6FB}} \color[HTML]{000000} 0.5045 \\
\hline
\multirow[c]{3}{*}{Detection-Longformer} & original & {\cellcolor[HTML]{F6EFF7}} \color[HTML]{000000} 0.5616 & {\cellcolor[HTML]{F0EAF4}} \color[HTML]{000000} 0.6032 & \bfseries {\cellcolor[HTML]{FFF7FB}} \color[HTML]{000000} 0.5010 & {\cellcolor[HTML]{EFE9F3}} \color[HTML]{000000} 0.6065 & {\cellcolor[HTML]{EDE8F3}} \color[HTML]{000000} 0.6179 & {\cellcolor[HTML]{EEE9F3}} \color[HTML]{000000} 0.6097 & {\cellcolor[HTML]{FFF7FB}} \color[HTML]{000000} 0.4760 & {\cellcolor[HTML]{F9F2F8}} \color[HTML]{000000} 0.5391 \\
 & paraphrased & {\cellcolor[HTML]{FEF6FA}} \color[HTML]{000000} 0.5107 & {\cellcolor[HTML]{FBF4F9}} \color[HTML]{000000} 0.5247 & {\cellcolor[HTML]{FFF7FB}} \color[HTML]{000000} 0.4216 & {\cellcolor[HTML]{F2ECF5}} \color[HTML]{000000} 0.5887 & {\cellcolor[HTML]{FBF3F9}} \color[HTML]{000000} 0.5287 & {\cellcolor[HTML]{F5EEF6}} \color[HTML]{000000} 0.5683 & {\cellcolor[HTML]{FFF7FB}} \color[HTML]{000000} 0.4253 & {\cellcolor[HTML]{FBF4F9}} \color[HTML]{000000} 0.5246 \\
 & homoglyph & \bfseries {\cellcolor[HTML]{EDE8F3}} \color[HTML]{000000} 0.6196 & \bfseries {\cellcolor[HTML]{EBE6F2}} \color[HTML]{000000} 0.6319 & {\cellcolor[HTML]{FFF7FB}} \color[HTML]{000000} 0.4883 & \bfseries {\cellcolor[HTML]{DFDDEC}} \color[HTML]{000000} 0.6870 & \bfseries {\cellcolor[HTML]{E7E3F0}} \color[HTML]{000000} 0.6492 & \bfseries {\cellcolor[HTML]{E7E3F0}} \color[HTML]{000000} 0.6504 & \bfseries {\cellcolor[HTML]{F7F0F7}} \color[HTML]{000000} 0.5537 & \bfseries {\cellcolor[HTML]{E0DDED}} \color[HTML]{000000} 0.6804 \\
\hline
\end{tabular}
}
\caption{Per-test-language comparison of performance (AUC ROC) of the selected MGT detectors on original and adversarial data. Bold represents the highest value per each test language and each detector (in regard to original, paraphrased, and homoglyph subset).}
\label{tab:adversarial}
\end{table*}

\begin{table*}[!t]
\centering
\resizebox{\textwidth}{!}{
\addtolength{\tabcolsep}{-2pt}
\begin{tabular}{c|l||c|c|c|c|c|c|c|c}
\hline
\textbf{Detector} & \textbf{Subset} & \bfseries All & \bfseries cs & \bfseries de & \bfseries hr & \bfseries hu & \bfseries pl & \bfseries sk & \bfseries sl \\
\hline
\multirow[c]{2}{*}{Llama-3.2-3B} & paraphrased & \bfseries {\cellcolor[HTML]{FFF7EC}} \color[HTML]{000000} 1.1478 & \bfseries {\cellcolor[HTML]{FFF7EC}} \color[HTML]{000000} 1.3756 & \bfseries {\cellcolor[HTML]{FFF7EC}} \color[HTML]{000000} 0.6972 & \bfseries {\cellcolor[HTML]{FFF7EC}} \color[HTML]{000000} 0.8615 & \bfseries {\cellcolor[HTML]{FFF7EC}} \color[HTML]{000000} 0.4465 & \bfseries {\cellcolor[HTML]{FFF7EC}} \color[HTML]{000000} 0.5965 & \bfseries {\cellcolor[HTML]{FFF7EC}} \color[HTML]{000000} 1.4596 & \bfseries {\cellcolor[HTML]{FFF7EC}} \color[HTML]{000000} 1.7855 \\
 & homoglyph & {\cellcolor[HTML]{FFF4E5}} \color[HTML]{000000} -3.9598 & {\cellcolor[HTML]{FFF6EA}} \color[HTML]{000000} -1.6017 & {\cellcolor[HTML]{FFF4E4}} \color[HTML]{000000} -4.4729 & {\cellcolor[HTML]{FFF6EA}} \color[HTML]{000000} -1.5661 & {\cellcolor[HTML]{FFF5E6}} \color[HTML]{000000} -3.2416 & {\cellcolor[HTML]{FFF3E3}} \color[HTML]{000000} -4.6977 & {\cellcolor[HTML]{FFF4E4}} \color[HTML]{000000} -4.3851 & {\cellcolor[HTML]{FFF3E3}} \color[HTML]{000000} -5.1506 \\
\hline
\multirow[c]{2}{*}{mDeBERTa-v3-base} & paraphrased & \bfseries {\cellcolor[HTML]{FFF7EC}} \color[HTML]{000000} 1.0807 & \bfseries {\cellcolor[HTML]{FFF7EC}} \color[HTML]{000000} 1.1490 & \bfseries {\cellcolor[HTML]{FFF7EC}} \color[HTML]{000000} 0.8103 & \bfseries {\cellcolor[HTML]{FFF7EC}} \color[HTML]{000000} 1.1151 & \bfseries {\cellcolor[HTML]{FFF7EC}} \color[HTML]{000000} 0.5387 & \bfseries {\cellcolor[HTML]{FFF7EC}} \color[HTML]{000000} 0.7277 & \bfseries {\cellcolor[HTML]{FFF7EC}} \color[HTML]{000000} 1.7101 & \bfseries {\cellcolor[HTML]{FFF7EC}} \color[HTML]{000000} 1.1182 \\
 & homoglyph & {\cellcolor[HTML]{FEE7C5}} \color[HTML]{000000} -20.2776 & {\cellcolor[HTML]{FEEDD4}} \color[HTML]{000000} -12.8004 & {\cellcolor[HTML]{FDDBAD}} \color[HTML]{000000} -30.8013 & {\cellcolor[HTML]{FEEACC}} \color[HTML]{000000} -16.4754 & {\cellcolor[HTML]{FEEACE}} \color[HTML]{000000} -16.3152 & {\cellcolor[HTML]{FEE2BC}} \color[HTML]{000000} -24.4904 & {\cellcolor[HTML]{FEEACE}} \color[HTML]{000000} -15.9545 & {\cellcolor[HTML]{FEDFB5}} \color[HTML]{000000} -27.1115 \\
\hline
\multirow[c]{2}{*}{Gemma-2-2B} & paraphrased & \bfseries {\cellcolor[HTML]{FFF7EC}} \color[HTML]{000000} 1.9915 & \bfseries {\cellcolor[HTML]{FFF7EC}} \color[HTML]{000000} 1.2283 & \bfseries {\cellcolor[HTML]{FFF7EC}} \color[HTML]{000000} 1.0638 & \bfseries {\cellcolor[HTML]{FFF7EC}} \color[HTML]{000000} 2.8481 & \bfseries {\cellcolor[HTML]{FFF7EC}} \color[HTML]{000000} 0.2619 & \bfseries {\cellcolor[HTML]{FFF7EC}} \color[HTML]{000000} 0.7575 & \bfseries {\cellcolor[HTML]{FFF7EC}} \color[HTML]{000000} 1.6004 & \bfseries {\cellcolor[HTML]{FFF7EC}} \color[HTML]{000000} 2.7112 \\
 & homoglyph & {\cellcolor[HTML]{FFF2E0}} \color[HTML]{000000} -6.5255 & {\cellcolor[HTML]{FFF4E5}} \color[HTML]{000000} -4.0138 & {\cellcolor[HTML]{FFF5E6}} \color[HTML]{000000} -3.3260 & {\cellcolor[HTML]{FFF2E1}} \color[HTML]{000000} -5.9583 & {\cellcolor[HTML]{FFF5E7}} \color[HTML]{000000} -2.5900 & {\cellcolor[HTML]{FFF2E0}} \color[HTML]{000000} -6.5343 & {\cellcolor[HTML]{FFF2E0}} \color[HTML]{000000} -6.8872 & {\cellcolor[HTML]{FEEFDA}} \color[HTML]{000000} -9.4358 \\
\hline
\multirow[c]{2}{*}{XLM-RoBERTa-base} & paraphrased & \bfseries {\cellcolor[HTML]{FFF7EC}} \color[HTML]{000000} 1.3034 & \bfseries {\cellcolor[HTML]{FFF7EC}} \color[HTML]{000000} 1.4131 & \bfseries {\cellcolor[HTML]{FFF7EC}} \color[HTML]{000000} 0.6518 & \bfseries {\cellcolor[HTML]{FFF7EC}} \color[HTML]{000000} 1.5217 & \bfseries {\cellcolor[HTML]{FFF7EC}} \color[HTML]{000000} 0.7100 & \bfseries {\cellcolor[HTML]{FFF7EC}} \color[HTML]{000000} 0.1951 & \bfseries {\cellcolor[HTML]{FFF7EC}} \color[HTML]{000000} 2.6804 & \bfseries {\cellcolor[HTML]{FFF7EC}} \color[HTML]{000000} 2.2970 \\
 & homoglyph & {\cellcolor[HTML]{FDCA93}} \color[HTML]{000000} -45.3002 & {\cellcolor[HTML]{FDD8A7}} \color[HTML]{000000} -33.6366 & {\cellcolor[HTML]{FDB27B}} \color[HTML]{000000} -60.0206 & {\cellcolor[HTML]{FDCE97}} \color[HTML]{000000} -42.7661 & {\cellcolor[HTML]{FDD8A6}} \color[HTML]{000000} -34.3776 & {\cellcolor[HTML]{FDCA94}} \color[HTML]{000000} -44.8501 & {\cellcolor[HTML]{FDC58E}} \color[HTML]{000000} -49.0268 & {\cellcolor[HTML]{FDBB85}} \color[HTML]{000000} -55.7142 \\
\hline
\multirow[c]{2}{*}{Fast-DetectGPT} & paraphrased & \bfseries {\cellcolor[HTML]{FEEED7}} \color[HTML]{000000} -11.5706 & \bfseries {\cellcolor[HTML]{FFF3E3}} \color[HTML]{000000} -4.8167 & \bfseries {\cellcolor[HTML]{FFF5E7}} \color[HTML]{000000} -2.4718 & \bfseries {\cellcolor[HTML]{FEEFD9}} \color[HTML]{000000} -10.1836 & \bfseries {\cellcolor[HTML]{FDD7A4}} \color[HTML]{000000} -35.0241 & \bfseries {\cellcolor[HTML]{FEECD2}} \color[HTML]{000000} -13.9639 & \bfseries {\cellcolor[HTML]{FFF4E4}} \color[HTML]{000000} -4.1148 & \bfseries {\cellcolor[HTML]{FFF0DB}} \color[HTML]{000000} -8.8522 \\
 & homoglyph & {\cellcolor[HTML]{F26D4B}} \color[HTML]{000000} -89.8062 & {\cellcolor[HTML]{F4754F}} \color[HTML]{000000} -86.4184 & {\cellcolor[HTML]{ED6145}} \color[HTML]{000000} -95.0409 & {\cellcolor[HTML]{EF6548}} \color[HTML]{000000} -93.2612 & {\cellcolor[HTML]{F26D4B}} \color[HTML]{000000} -90.0897 & {\cellcolor[HTML]{F26E4C}} \color[HTML]{000000} -89.4144 & {\cellcolor[HTML]{F67A51}} \color[HTML]{000000} -83.9193 & {\cellcolor[HTML]{F16C4B}} \color[HTML]{000000} -90.7533 \\
\hline
\multirow[c]{2}{*}{Binoculars} & paraphrased & \bfseries {\cellcolor[HTML]{FFF0DC}} \color[HTML]{000000} -8.5022 & \bfseries {\cellcolor[HTML]{FFF5E7}} \color[HTML]{000000} -2.6775 & \bfseries {\cellcolor[HTML]{FFF7EB}} \color[HTML]{000000} -0.7753 & \bfseries {\cellcolor[HTML]{FFF1DE}} \color[HTML]{000000} -7.3183 & \bfseries {\cellcolor[HTML]{FEE0B8}} \color[HTML]{000000} -26.3440 & \bfseries {\cellcolor[HTML]{FEEFD9}} \color[HTML]{000000} -10.0776 & \bfseries {\cellcolor[HTML]{FFF6E9}} \color[HTML]{000000} -2.1252 & \bfseries {\cellcolor[HTML]{FFF2E1}} \color[HTML]{000000} -6.3132 \\
 & homoglyph & {\cellcolor[HTML]{FDA56F}} \color[HTML]{000000} -65.0519 & {\cellcolor[HTML]{FDAC76}} \color[HTML]{000000} -62.6195 & {\cellcolor[HTML]{FC9863}} \color[HTML]{000000} -70.4672 & {\cellcolor[HTML]{FC9964}} \color[HTML]{000000} -70.1066 & {\cellcolor[HTML]{FDB37D}} \color[HTML]{000000} -59.7577 & {\cellcolor[HTML]{FDA872}} \color[HTML]{000000} -64.1452 & {\cellcolor[HTML]{FDB37D}} \color[HTML]{000000} -59.3729 & {\cellcolor[HTML]{FC9863}} \color[HTML]{000000} -70.8711 \\
\hline
\multirow[c]{2}{*}{LLM-Deviation} & paraphrased & \bfseries {\cellcolor[HTML]{FFF5E7}} \color[HTML]{000000} -2.7774 & \bfseries {\cellcolor[HTML]{FFF6E9}} \color[HTML]{000000} -2.0273 & \bfseries {\cellcolor[HTML]{FFF7EC}} \color[HTML]{000000} 1.2324 & \bfseries {\cellcolor[HTML]{FFF7EB}} \color[HTML]{000000} -0.9388 & \bfseries {\cellcolor[HTML]{FEE9CA}} \color[HTML]{000000} -17.9738 & \bfseries {\cellcolor[HTML]{FFF1DD}} \color[HTML]{000000} -8.1312 & \bfseries {\cellcolor[HTML]{FFF7EC}} \color[HTML]{000000} 0.6255 & \bfseries {\cellcolor[HTML]{FFF6E9}} \color[HTML]{000000} -2.1857 \\
 & homoglyph & {\cellcolor[HTML]{FDC68F}} \color[HTML]{000000} -48.1313 & {\cellcolor[HTML]{FDCC96}} \color[HTML]{000000} -43.5384 & {\cellcolor[HTML]{FDA671}} \color[HTML]{000000} -64.7306 & {\cellcolor[HTML]{FDC48D}} \color[HTML]{000000} -49.6675 & {\cellcolor[HTML]{FDCB95}} \color[HTML]{000000} -43.9930 & {\cellcolor[HTML]{FDC28B}} \color[HTML]{000000} -51.0777 & {\cellcolor[HTML]{FDC28B}} \color[HTML]{000000} -51.2818 & {\cellcolor[HTML]{FDBC85}} \color[HTML]{000000} -55.4268 \\
\hline
\multirow[c]{2}{*}{BLOOMz-3B-mixed-detector} & paraphrased & \bfseries \bfseries {\cellcolor[HTML]{FFF7EC}} \color[HTML]{000000} 25.2145 & \bfseries \bfseries {\cellcolor[HTML]{FFF7EC}} \color[HTML]{000000} 29.8580 & \bfseries \bfseries {\cellcolor[HTML]{FFF7EC}} \color[HTML]{000000} 24.9264 & \bfseries \bfseries {\cellcolor[HTML]{FFF7EC}} \color[HTML]{000000} 21.5639 & \bfseries \bfseries {\cellcolor[HTML]{FFF7EC}} \color[HTML]{000000} 20.4258 & \bfseries \bfseries {\cellcolor[HTML]{FFF7EC}} \color[HTML]{000000} 27.2610 & \bfseries \bfseries {\cellcolor[HTML]{FFF7EC}} \color[HTML]{000000} 25.5008 & \bfseries \bfseries {\cellcolor[HTML]{FFF7EC}} \color[HTML]{000000} 28.2813 \\
 & homoglyph & {\cellcolor[HTML]{FDD49F}} \color[HTML]{000000} -37.4430 & {\cellcolor[HTML]{FDD8A6}} \color[HTML]{000000} -33.9972 & {\cellcolor[HTML]{FDD49F}} \color[HTML]{000000} -37.3355 & {\cellcolor[HTML]{FDD29C}} \color[HTML]{000000} -38.9864 & {\cellcolor[HTML]{FDD39D}} \color[HTML]{000000} -38.1557 & {\cellcolor[HTML]{FDD5A0}} \color[HTML]{000000} -36.4922 & {\cellcolor[HTML]{FDD6A3}} \color[HTML]{000000} -35.7036 & {\cellcolor[HTML]{FDCC96}} \color[HTML]{000000} -43.7688 \\
\hline
\multirow[c]{2}{*}{ChatGPT-detector-RoBERTa-Chinese} & paraphrased & \bfseries {\cellcolor[HTML]{FFF7EC}} \color[HTML]{000000} 0.2833 & \bfseries {\cellcolor[HTML]{FFF4E5}} \color[HTML]{000000} -3.5264 & \bfseries {\cellcolor[HTML]{FFF6EA}} \color[HTML]{000000} -1.2455 & \bfseries {\cellcolor[HTML]{FFF7EC}} \color[HTML]{000000} 4.1424 & \bfseries {\cellcolor[HTML]{FFF7EC}} \color[HTML]{000000} 2.4885 & \bfseries {\cellcolor[HTML]{FFF6EA}} \color[HTML]{000000} -1.5669 & \bfseries {\cellcolor[HTML]{FFF7EC}} \color[HTML]{000000} 4.5022 & \bfseries {\cellcolor[HTML]{FFF1DE}} \color[HTML]{000000} -7.2485 \\
 & homoglyph & {\cellcolor[HTML]{FEECD1}} \color[HTML]{000000} -14.6194 & {\cellcolor[HTML]{FEE9CA}} \color[HTML]{000000} -17.6742 & {\cellcolor[HTML]{FEE5C1}} \color[HTML]{000000} -21.8579 & {\cellcolor[HTML]{FEE7C7}} \color[HTML]{000000} -19.9088 & {\cellcolor[HTML]{FEECD2}} \color[HTML]{000000} -13.9314 & {\cellcolor[HTML]{FEEDD3}} \color[HTML]{000000} -13.3594 & {\cellcolor[HTML]{FEE9CB}} \color[HTML]{000000} -17.1165 & {\cellcolor[HTML]{FEE1B9}} \color[HTML]{000000} -25.4668 \\
\hline
\multirow[c]{2}{*}{Detection-Longformer} & paraphrased & {\cellcolor[HTML]{FFF0DB}} \color[HTML]{000000} -9.0621 & {\cellcolor[HTML]{FEEDD3}} \color[HTML]{000000} -13.0217 & {\cellcolor[HTML]{FEEACE}} \color[HTML]{000000} -15.8487 & {\cellcolor[HTML]{FFF5E6}} \color[HTML]{000000} -2.9369 & {\cellcolor[HTML]{FEECD1}} \color[HTML]{000000} -14.4351 & {\cellcolor[HTML]{FFF2E0}} \color[HTML]{000000} -6.7924 & {\cellcolor[HTML]{FEEFD8}} \color[HTML]{000000} -10.6439 & {\cellcolor[HTML]{FFF5E7}} \color[HTML]{000000} -2.6852 \\
 & homoglyph & \bfseries {\cellcolor[HTML]{FFF7EC}} \color[HTML]{000000} 10.3109 & \bfseries {\cellcolor[HTML]{FFF7EC}} \color[HTML]{000000} 4.7557 & \bfseries {\cellcolor[HTML]{FFF5E7}} \color[HTML]{000000} -2.5400 & \bfseries {\cellcolor[HTML]{FFF7EC}} \color[HTML]{000000} 13.2644 & \bfseries {\cellcolor[HTML]{FFF7EC}} \color[HTML]{000000} 5.0531 & \bfseries {\cellcolor[HTML]{FFF7EC}} \color[HTML]{000000} 6.6735 & \bfseries {\cellcolor[HTML]{FFF7EC}} \color[HTML]{000000} 16.3349 & \bfseries {\cellcolor[HTML]{FFF7EC}} \color[HTML]{000000} 26.2238 \\
\hline
\end{tabular}
}
\caption{Per-test-language evaluation of adversarial robustness of the selected MGT detectors as a difference in performance (AUC ROC) on the obfuscated adversarial data and on the original data. Bold represents the highest value per each test language and each detector (in regard to individual obfuscated subsets).}
\label{tab:adversarial_robustness}
\end{table*}

To compare the performance of MGT detectors for each of the two domains (news and social media), we are providing the per-language comparison for each domain separately in Table~\ref{tab:domains}, analogously to Table~\ref{tab:categories}. We can observe that the \textbf{finetuned detectors dominate in both domains}, while news articles (longer and more formal texts) are a bit easier for the finetuned detectors. Interestingly, none of the train-languages combination of the best-performing finetuned models is the same across the two domains; therefore, there is none clearly dominating combination. Furthermore, for mDeBERTa base model, a single-train language (Czech for News, German for Social media) achieved the best performance. Polish and Slovenian social-media texts are the most difficult for detection. In other categories, LLM-Deviation and BLOOMZ-3B-mixed-detector are the worst in the news domain, while being the best in the social-media domain. Detection-Longformer has similarly the better performance in news, while performing worse than random classifier in the social-media domain. For pretrained detectors, it definitely depends on the domain of their pretraining data. But, the case is different for LLM-Deviation, since it is zero-shot statistical detection metric (i.e., without training) and this effect is not consistent among languages (most significant in Hungarian and Polish).

\subsection{Adversarial Robustness Evaluation}

We have evaluated adversarial robustness of the MGT detectors against obfuscation by paraphrasing and by homoglyph attack. The results, provided in Table~\ref{tab:adversarial} and Table~\ref{tab:adversarial_robustness}, indicate that \textbf{finetuned detectors are the most robust towards obfuscation} (especially bigger ones). With a sole exception of Detection-Longformer (where the performance is actually increased), the \textbf{homoglyph-based obfuscation decreases the performance significantly more than paraphrasing} (in all languages). Detection-Longformer is more susceptible to paraphrasing. The other two pretrained detectors and finetuned detectors are mostly immune to this kind of obfuscation across all the languages. Statistical detectors are confused by both obfuscation methods, while homoglyph attack can decrease the AUC ROC performance by 95\% (in case of Fast-DetectGPT for German texts).

\section{Conclusions}
\label{sec:conclusion}

This benchmark study, focused on a set of Central European languages, brought several insights in the detection of machine-generated texts for a bunch of under-researched languages. The comparison of the effect of language combinations on detectors finetuning revealed only small differences in generalization towards other Central European languages. Out of the compared statistical, pretrained, and finetuned categories of MGT detectors, the last one is performing significantly better across all the tested languages. The finetuned detectors are also the most robust against paraphrasing and homoglyph-based obfuscation, making them the most suitable for Central European languages. However, until now there have been only few models available in some of these languages. This further signifies the need to perform research also in the languages left usually out-of-focus of the mainstream.

\section*{Limitations}
\label{sec:limitations}
Although this study covers 7 languages, 2 domains, 8 LLMs, 2 authorship obfuscation techniques, and 10 MGT detection methods, the results might still be biased based on some specific aspects of the data, affecting the generalizability of the conclusions to out-of-distribution data (e.g., other languages, other LLM generators). For example, the selected languages cover 3 language-family branches, but 5 of the languages are Slavic. This is, however, specific bias of the Central European geographic area, which was the aim of this study. We have done a manual hyper-parameters optimization of the detectors' finetuning process; however, we have covered just a small set of options, where a further tuning might increase the performance and generalizability of the individual detectors.

\section*{Ethics Statement}
\label{sec:ethics}
Our work is focused on evaluation and comparison (i.e., benchmarking) of the MGT detection methods on 7 languages of Central European region, bringing important insights in these under-researched languages. We use the existing datasets in our work in accordance with their intended use and licenses (for research purpose only). As a part of our work, we are not re-sharing any existing data or publishing any new dataset. For the research replicability and validation purposes, we are publishing the pre-processing, training, and evaluation source codes (for research purposes only). The existing artifacts used in this work have been properly cited and used according their licenses and intended use. We have also checked and followed licensing and terms of use of the used LLMs. AI assistants have not been used for conducting research in any other way than already described in the paper (text obfuscation, finetuning and detection of MGTs).

\section*{Acknowledgments}
Funded by the EU NextGenerationEU through the Recovery and Resilience Plan for Slovakia under the project No. 09I01-03-V04-00059.

\textbf{Computational resources}. Part of the research results was obtained using the computational resources procured in the national project \textit{National competence centre for high performance computing} (project code: 311070AKF2) funded by European Regional Development Fund, EU Structural Funds Informatization of Society, Operational Program Integrated Infrastructure.

\bibliography{anthology,custom}

\appendix

\section{Computational Resources}
\label{sec:resources}

For finetuning and inference of machine-generated text detection models, we have used 1x A100 40GB, consuming about 2000 GPU-hours. For executing authorship obfuscation in machine-generated texts and subsequent text-quality analysis, we have used 1x A100 40GB, consuming about 70 GPU-hours. For other tasks, we have not used GPU acceleration.

\section{Data Analysis}
\label{sec:dataanalysis}

We have analyzed basic stylometric characteristics of the selected combined dataset in a form of character counts and word counts per each included generator and domain. The results of such analysis are summarized in Table~\ref{tab:data_stats}. Aya generated the longest news articles and Gemini generated the longest social-media texts. On the other hand, OPT-IML generated the shortest texts in both domains.

\begin{table*}[!t]
\centering
\resizebox{0.8\linewidth}{!}{
\addtolength{\tabcolsep}{-2pt}
\begin{tabular}{l|c|c|c|c}
\hline
 & \multicolumn{2}{c|}{\textbf{News}} & \multicolumn{2}{c}{\textbf{Social media}} \\
\textbf{Generator} & \textbf{CC} & \textbf{WC} & \textbf{CC} & \textbf{WC} \\
\hline
Human & 795.99 (±269.48) & 115.06 (±38.77) & 66.86 (±92.36) & 9.09 (±11.16) \\
Llama-2-70B-Chat-HF & 986.08 (±368.97) & 147.24 (±57.12) & - & - \\
Mistral-7B-Instruct-v0.2 & 965.22 (±305.23) & 137.08 (±44.46) & 107.82 (±74.68) & 15.03 (±9.57) \\
Aya-101 & 1069.99 (±420.08) & 163.41 (±65.00) & 60.12 (±60.09) & 9.65 (±8.73) \\
Gemini & - & - & 435.09 (±365.40) & 65.68 (±50.87) \\
GPT-3.5-Turbo-0125 & 742.37 (±260.78) & 105.05 (±36.25) & 100.38 (±99.01) & 16.67 (±16.85) \\
OPT-IML-Max-30B & 491.70 (±353.00) & 74.47 (±42.75) & 47.02 (±44.46) & 7.43 (±6.57) \\
v5-Eagle-7B-HF & 923.74 (±347.38) & 135.26 (±51.95) & 111.15 (±85.21) & 16.87 (±12.35) \\
Vicuna-13B & 840.92 (±343.39) & 124.90 (±51.55) & 93.15 (±77.87) & 14.16 (±11.11) \\
\hline
\end{tabular}
}
\caption{Dataset statistics of mean count (+- standard deviation) of characters (CC) and words (WC) per generator and per domain.}
\label{tab:data_stats}
\end{table*}

\begin{figure}[!t]
\centering
\includegraphics[width=\linewidth]{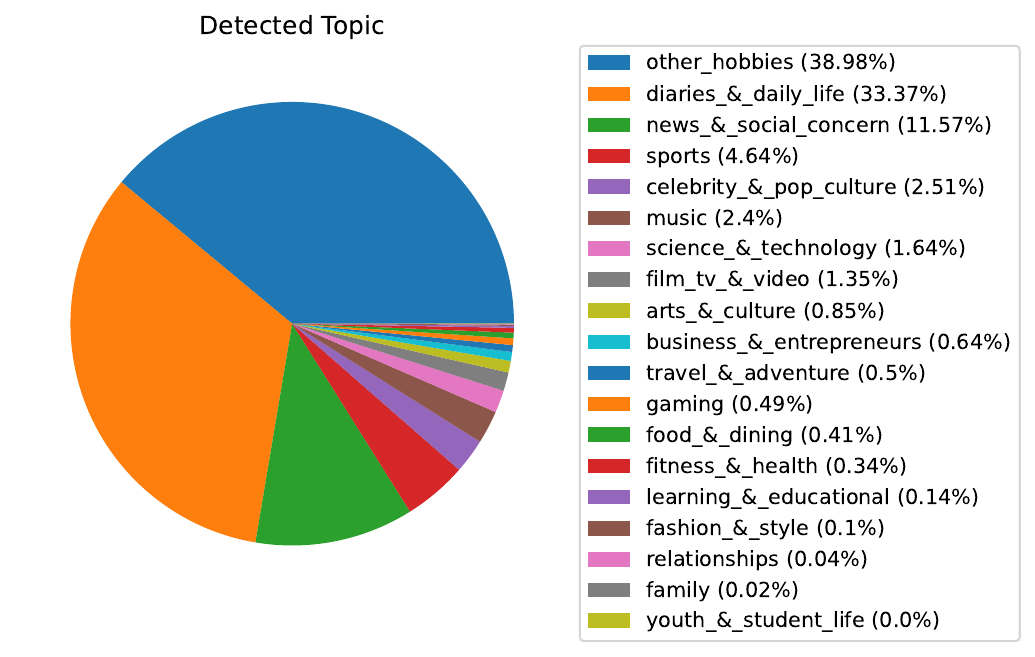}
\caption{Detected topics in the selected dataset.}
\label{fig:topics}
\end{figure}

\begin{figure}[!t]
\centering
\includegraphics[width=\linewidth]{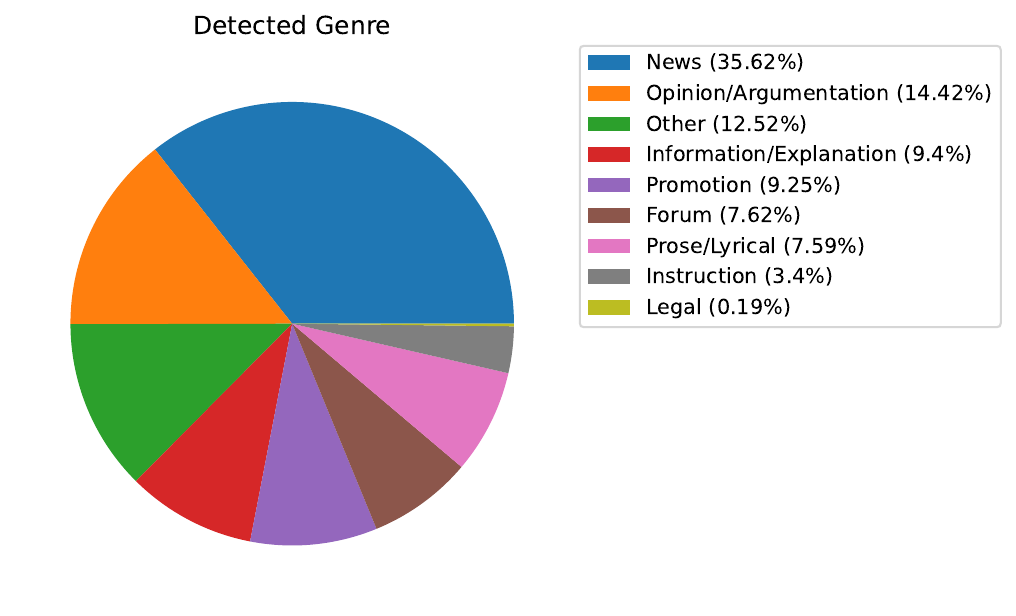}
\caption{Detected genres in the selected dataset.}
\label{fig:genres}
\end{figure}

Similarly to the MultiSocial study, we have used analysis of topics and genres based on existing available detectors. The overview of the results of the topic detector\footnote{\scriptsize\url{https://huggingface.co/cardiffnlp/tweet-topic-latest-multi}}~\citep{antypas-etal-2022-twitter} is illustrated in \figurename~\ref{fig:topics}. The results overview of the multilingual text genre detector\footnote{\scriptsize\url{https://huggingface.co/classla/xlm-roberta-base-multilingual-text-genre-classifier}}~\citep{kuzman2023automatic}is illustrated in \figurename~\ref{fig:genres}. Although the texts are combined from news and social-media domains, there is variety of topics and genres in the data, making the results of this study representative (limiting the topical bias).


After application of the two authorship obfuscation techniques on subset of data, we have run a similarity analysis using various standard metrics, defined by \citealp{macko-etal-2024-authorship}, to compare original and obfuscated texts (Table~\ref{tab:adversarial_similarity}). It seems that paraphrasing significantly prolonged the texts, which might indicate an easier subsequent detectability. The semantic similarity indicated by BERTScore seems to be high after both obfuscations. Language detection seems to be not very accurate in shorter social-media texts.

\begin{table}[!t]
\centering
\resizebox{0.8\linewidth}{!}{
\addtolength{\tabcolsep}{-2pt}
\begin{tabular}{l|c|c}
\hline
 & \bfseries Paraphrased & \bfseries Homoglyph \\
\hline
\bfseries METEOR & 0.477 (±0.23) & 0.216 (±0.24) \\
\bfseries BERTScore & 0.824 (±0.09) & 0.872 (±0.05) \\
\bfseries ngram & 0.429 (±0.20) & 0.619 (±0.08) \\
\bfseries TF & 0.684 (±0.22) & 0.311 (±0.25) \\
\bfseries LD & 1.489 (±16.15) & 0.099 (±0.03) \\
\bfseries CharLenDiff & 1.988 (±16.18) & 1.061 (±0.35) \\
\bfseries LangCheck & 17.64\% & 14.64\% \\
\hline
\end{tabular}
}
\caption{Similarity of obfuscated texts to the original.}
\label{tab:adversarial_similarity}
\end{table}

\section{Results Data}
\label{sec:additionalresults}

Table~\ref{tab:trainlang_tpr}, Table~\ref{tab:categories_tpr}, and Table~\ref{tab:adversarial_tpr} contain the performance comparison using the TPR @ 5\% FPR metric, reflecting expected performance in the real world (where FPR must be minimized). These results indicate that even the finetuned detectors are far from perfect and the performance must be further tuned in the future.

\begin{table*}[!t]
\centering
\resizebox{\textwidth}{!}{
\addtolength{\tabcolsep}{-2pt}
\begin{tabular}{l|cc|cc|cc|cc|cc|cc|cc|cc}
\hline
\textbf{Train} & \multicolumn{2}{c|}{\textbf{All}} & \multicolumn{2}{c|}{\textbf{cs}} & \multicolumn{2}{c|}{\textbf{de}} & \multicolumn{2}{c|}{\textbf{hr}} & \multicolumn{2}{c|}{\textbf{hu}} & \multicolumn{2}{c|}{\textbf{pl}} & \multicolumn{2}{c|}{\textbf{sk}} & \multicolumn{2}{c}{\textbf{sl}} \\
\cline{2-17}
\textbf{Languages} & mean & std & mean & std & mean & std & mean & std & mean & std & mean & std & mean & std & mean & std \\
\hline
cs-de-hr-hu-pl & \bfseries 0.849 & 0.036 & 0.936 & 0.019 & 0.844 & 0.080 & 0.893 & 0.020 & 0.922 & 0.038 & 0.825 & 0.054 & 0.863 & 0.041 & 0.275 & 0.337 \\
de-pl & 0.844 & 0.033 & 0.916 & 0.023 & 0.863 & 0.045 & 0.850 & 0.065 & 0.901 & 0.050 & 0.849 & 0.062 & 0.845 & 0.056 & 0.731 & 0.033 \\
cs-de & 0.840 & 0.037 & 0.937 & 0.019 & \bfseries 0.879 & 0.039 & 0.855 & 0.051 & 0.893 & 0.047 & 0.762 & 0.086 & 0.855 & 0.045 & \bfseries 0.750 & 0.067 \\
cs-de-pl & 0.837 & 0.037 & 0.932 & 0.021 & 0.843 & 0.056 & 0.820 & 0.086 & 0.909 & 0.032 & 0.837 & 0.057 & 0.858 & 0.052 & 0.626 & 0.258 \\
de-hu-pl & 0.833 & 0.046 & 0.920 & 0.039 & 0.840 & 0.066 & 0.846 & 0.059 & 0.929 & 0.044 & 0.829 & 0.074 & 0.849 & 0.039 & 0.746 & 0.074 \\
cs-de-hr-pl & 0.832 & 0.033 & 0.928 & 0.018 & 0.850 & 0.043 & 0.886 & 0.027 & 0.865 & 0.039 & 0.817 & 0.051 & 0.821 & 0.057 & 0.528 & 0.356 \\
de-hr-pl & 0.822 & 0.045 & 0.928 & 0.028 & 0.838 & 0.047 & 0.889 & 0.049 & 0.897 & 0.031 & 0.804 & 0.095 & \bfseries 0.886 & 0.029 & 0.351 & 0.405 \\
cs-hu-pl & 0.813 & 0.061 & \bfseries 0.941 & 0.033 & 0.572 & 0.387 & 0.868 & 0.038 & \bfseries 0.951 & 0.020 & 0.845 & 0.061 & 0.820 & 0.101 & 0.650 & 0.100 \\
de-hu & 0.810 & 0.046 & 0.922 & 0.017 & 0.868 & 0.040 & 0.815 & 0.071 & 0.942 & 0.021 & 0.736 & 0.053 & 0.832 & 0.022 & 0.699 & 0.032 \\
cs-de-hr & 0.802 & 0.060 & 0.933 & 0.030 & 0.851 & 0.037 & 0.877 & 0.033 & 0.899 & 0.021 & 0.575 & 0.383 & 0.824 & 0.061 & 0.002 & 0.003 \\
cs-hr-pl & 0.801 & 0.045 & 0.934 & 0.031 & 0.521 & 0.369 & 0.890 & 0.014 & 0.906 & 0.035 & 0.837 & 0.054 & 0.821 & 0.061 & 0.336 & 0.394 \\
cs-de-hu-pl & 0.797 & 0.067 & 0.915 & 0.046 & 0.841 & 0.077 & 0.674 & 0.367 & 0.905 & 0.046 & 0.812 & 0.059 & 0.826 & 0.081 & 0.569 & 0.304 \\
hu-pl & 0.793 & 0.047 & 0.926 & 0.038 & 0.308 & 0.360 & 0.822 & 0.061 & 0.950 & 0.020 & 0.599 & 0.403 & 0.785 & 0.072 & 0.466 & 0.313 \\
cs-de-hu & 0.783 & 0.061 & 0.897 & 0.094 & 0.852 & 0.073 & 0.770 & 0.109 & 0.923 & 0.038 & 0.477 & 0.380 & 0.596 & 0.402 & 0.397 & 0.330 \\
cs-hr-hu & 0.782 & 0.104 & 0.916 & 0.068 & 0.742 & 0.162 & 0.867 & 0.046 & 0.926 & 0.070 & 0.689 & 0.149 & 0.785 & 0.098 & 0.263 & 0.307 \\
cs & 0.779 & 0.054 & 0.928 & 0.011 & 0.595 & 0.397 & 0.810 & 0.077 & 0.877 & 0.075 & 0.403 & 0.466 & 0.631 & 0.422 & 0.679 & 0.056 \\
de & 0.776 & 0.074 & 0.879 & 0.044 & 0.875 & 0.034 & 0.767 & 0.102 & 0.816 & 0.129 & 0.719 & 0.058 & 0.825 & 0.064 & 0.709 & 0.063 \\
pl & 0.772 & 0.119 & 0.907 & 0.084 & 0.427 & 0.493 & 0.849 & 0.062 & 0.895 & 0.068 & \bfseries 0.858 & 0.038 & 0.820 & 0.127 & 0.710 & 0.083 \\
cs-de-hr-hu & 0.769 & 0.097 & 0.927 & 0.030 & 0.860 & 0.060 & \bfseries 0.903 & 0.025 & 0.935 & 0.031 & 0.751 & 0.082 & 0.570 & 0.405 & 0.171 & 0.341 \\
cs-pl & 0.740 & 0.133 & 0.900 & 0.107 & 0.380 & 0.443 & 0.799 & 0.088 & 0.876 & 0.073 & 0.810 & 0.118 & 0.799 & 0.116 & 0.512 & 0.341 \\
cs-hu & 0.686 & 0.194 & 0.913 & 0.042 & 0.644 & 0.225 & 0.828 & 0.038 & 0.932 & 0.030 & 0.662 & 0.139 & 0.673 & 0.206 & 0.514 & 0.120 \\
hu & 0.664 & 0.113 & 0.855 & 0.082 & 0.649 & 0.165 & 0.745 & 0.120 & 0.933 & 0.046 & 0.654 & 0.189 & 0.690 & 0.090 & 0.582 & 0.143 \\
de-hr-hu-pl & 0.628 & 0.422 & 0.927 & 0.029 & 0.851 & 0.065 & 0.891 & 0.016 & 0.689 & 0.460 & 0.618 & 0.420 & 0.627 & 0.419 & 0.189 & 0.378 \\
hr-hu & 0.599 & 0.401 & 0.906 & 0.057 & 0.575 & 0.384 & 0.865 & 0.039 & 0.934 & 0.021 & 0.476 & 0.356 & 0.411 & 0.475 & 0.206 & 0.412 \\
cs-hr-hu-pl & 0.596 & 0.400 & 0.937 & 0.024 & 0.535 & 0.374 & 0.877 & 0.023 & 0.946 & 0.030 & 0.817 & 0.057 & 0.573 & 0.385 & 0.171 & 0.341 \\
hr-hu-pl & 0.594 & 0.398 & 0.931 & 0.039 & 0.349 & 0.410 & 0.643 & 0.430 & 0.944 & 0.022 & 0.825 & 0.075 & 0.567 & 0.392 & 0.172 & 0.339 \\
de-hr & 0.594 & 0.400 & 0.918 & 0.028 & 0.875 & 0.042 & 0.902 & 0.007 & 0.900 & 0.040 & 0.628 & 0.264 & 0.840 & 0.035 & 0.072 & 0.145 \\
de-hr-hu & 0.555 & 0.383 & 0.899 & 0.059 & 0.822 & 0.087 & 0.877 & 0.029 & 0.936 & 0.015 & 0.731 & 0.075 & 0.567 & 0.406 & 0.162 & 0.324 \\
hr-pl & 0.391 & 0.454 & 0.931 & 0.033 & 0.550 & 0.377 & 0.901 & 0.019 & 0.910 & 0.029 & 0.834 & 0.077 & 0.782 & 0.116 & 0.192 & 0.383 \\
hr & 0.345 & 0.403 & 0.893 & 0.055 & 0.535 & 0.366 & 0.877 & 0.044 & 0.843 & 0.100 & 0.438 & 0.347 & 0.446 & 0.442 & 0.068 & 0.135 \\
cs-hr & 0.168 & 0.335 & 0.914 & 0.039 & 0.372 & 0.442 & 0.854 & 0.040 & 0.885 & 0.050 & 0.399 & 0.462 & 0.589 & 0.404 & 0.186 & 0.372 \\
\hline
\end{tabular}
}
\caption{Per-test-language comparison of performance (TPR @ 5\% FPR averaged across the finetuned base models) of finetuned MGT detectors based on combination of train languages. Bold represents the highest value per each test language.}
\label{tab:trainlang_tpr}
\end{table*}

\begin{table*}[!t]
\centering
\resizebox{\textwidth}{!}{
\addtolength{\tabcolsep}{-2pt}
\begin{tabular}{c|l||c|c|c|c|c|c|c|c}
\hline
\textbf{Category} & \textbf{Detector} & \bfseries All & \bfseries cs & \bfseries de & \bfseries hr & \bfseries hu & \bfseries pl & \bfseries sk & \bfseries sl \\
\hline
F & Llama-3.2-3B (pl) & \bfseries {\cellcolor[HTML]{9EBAD9}} \color[HTML]{000000} 0.8954 & {\cellcolor[HTML]{89B1D4}} \color[HTML]{000000} 0.9480 & \bfseries {\cellcolor[HTML]{A5BDDB}} \color[HTML]{000000} 0.8780 & \bfseries {\cellcolor[HTML]{91B5D6}} \color[HTML]{000000} 0.9260 & \bfseries {\cellcolor[HTML]{88B1D4}} \color[HTML]{000000} 0.9500 & \bfseries {\cellcolor[HTML]{9FBAD9}} \color[HTML]{000000} 0.8940 & \bfseries {\cellcolor[HTML]{8FB4D6}} \color[HTML]{000000} 0.9300 & {\cellcolor[HTML]{D0D1E6}} \color[HTML]{000000} 0.7520 \\
F & mDeBERTa-v3-base (de-pl-hr-hu) & {\cellcolor[HTML]{A7BDDB}} \color[HTML]{000000} 0.8749 & {\cellcolor[HTML]{8FB4D6}} \color[HTML]{000000} 0.9300 & \bfseries {\cellcolor[HTML]{A5BDDB}} \color[HTML]{000000} 0.8780 & {\cellcolor[HTML]{9EBAD9}} \color[HTML]{000000} 0.8980 & {\cellcolor[HTML]{94B6D7}} \color[HTML]{000000} 0.9180 & {\cellcolor[HTML]{A1BBDA}} \color[HTML]{000000} 0.8900 & {\cellcolor[HTML]{B1C2DE}} \color[HTML]{000000} 0.8400 & \bfseries {\cellcolor[HTML]{CED0E6}} \color[HTML]{000000} 0.7560 \\
F & Gemma-2-2B (de-pl-hr-hu) & {\cellcolor[HTML]{A7BDDB}} \color[HTML]{000000} 0.8731 & \bfseries {\cellcolor[HTML]{88B1D4}} \color[HTML]{000000} 0.9500 & \bfseries {\cellcolor[HTML]{A5BDDB}} \color[HTML]{000000} 0.8780 & {\cellcolor[HTML]{99B8D8}} \color[HTML]{000000} 0.9080 & {\cellcolor[HTML]{8CB3D5}} \color[HTML]{000000} 0.9380 & {\cellcolor[HTML]{A5BDDB}} \color[HTML]{000000} 0.8760 & {\cellcolor[HTML]{A7BDDB}} \color[HTML]{000000} 0.8720 & {\cellcolor[HTML]{FFF7FB}} \color[HTML]{000000} 0.0000 \\
F & XLM-RoBERTa-base (de-cs) & {\cellcolor[HTML]{B8C6E0}} \color[HTML]{000000} 0.8206 & {\cellcolor[HTML]{91B5D6}} \color[HTML]{000000} 0.9280 & {\cellcolor[HTML]{B9C6E0}} \color[HTML]{000000} 0.8200 & {\cellcolor[HTML]{ADC1DD}} \color[HTML]{000000} 0.8520 & {\cellcolor[HTML]{B7C5DF}} \color[HTML]{000000} 0.8260 & {\cellcolor[HTML]{D4D4E8}} \color[HTML]{000000} 0.7340 & {\cellcolor[HTML]{B7C5DF}} \color[HTML]{000000} 0.8260 & \bfseries {\cellcolor[HTML]{CED0E6}} \color[HTML]{000000} 0.7560 \\
S & Fast-DetectGPT & {\cellcolor[HTML]{FFF7FB}} \color[HTML]{000000} 0.4089 & {\cellcolor[HTML]{FFF7FB}} \color[HTML]{000000} 0.3980 & {\cellcolor[HTML]{FDF5FA}} \color[HTML]{000000} 0.5120 & {\cellcolor[HTML]{FFF7FB}} \color[HTML]{000000} 0.4580 & {\cellcolor[HTML]{FFF7FB}} \color[HTML]{000000} 0.4280 & {\cellcolor[HTML]{FFF7FB}} \color[HTML]{000000} 0.4520 & {\cellcolor[HTML]{FFF7FB}} \color[HTML]{000000} 0.3720 & {\cellcolor[HTML]{FFF7FB}} \color[HTML]{000000} 0.4140 \\
S & Binoculars & {\cellcolor[HTML]{FFF7FB}} \color[HTML]{000000} 0.3871 & {\cellcolor[HTML]{FFF7FB}} \color[HTML]{000000} 0.4740 & {\cellcolor[HTML]{FEF6FB}} \color[HTML]{000000} 0.5040 & {\cellcolor[HTML]{FEF6FA}} \color[HTML]{000000} 0.5100 & {\cellcolor[HTML]{FFF7FB}} \color[HTML]{000000} 0.4240 & {\cellcolor[HTML]{FFF7FB}} \color[HTML]{000000} 0.4200 & {\cellcolor[HTML]{FFF7FB}} \color[HTML]{000000} 0.3520 & {\cellcolor[HTML]{FFF7FB}} \color[HTML]{000000} 0.4160 \\
S & LLM-Deviation & {\cellcolor[HTML]{FFF7FB}} \color[HTML]{000000} 0.1806 & {\cellcolor[HTML]{FFF7FB}} \color[HTML]{000000} 0.3880 & {\cellcolor[HTML]{FFF7FB}} \color[HTML]{000000} 0.2560 & {\cellcolor[HTML]{FFF7FB}} \color[HTML]{000000} 0.4300 & {\cellcolor[HTML]{FFF7FB}} \color[HTML]{000000} 0.2020 & {\cellcolor[HTML]{FFF7FB}} \color[HTML]{000000} 0.3040 & {\cellcolor[HTML]{FFF7FB}} \color[HTML]{000000} 0.2540 & {\cellcolor[HTML]{FFF7FB}} \color[HTML]{000000} 0.4280 \\
P & BLOOMZ-3B-mixed-detector & {\cellcolor[HTML]{FFF7FB}} \color[HTML]{000000} 0.1540 & {\cellcolor[HTML]{FFF7FB}} \color[HTML]{000000} 0.1840 & {\cellcolor[HTML]{FFF7FB}} \color[HTML]{000000} 0.2140 & {\cellcolor[HTML]{FFF7FB}} \color[HTML]{000000} 0.1500 & {\cellcolor[HTML]{FFF7FB}} \color[HTML]{000000} 0.1920 & {\cellcolor[HTML]{FFF7FB}} \color[HTML]{000000} 0.1600 & {\cellcolor[HTML]{FFF7FB}} \color[HTML]{000000} 0.2380 & {\cellcolor[HTML]{FFF7FB}} \color[HTML]{000000} 0.0000 \\
P & ChatGPT-detector-RoBERTa-Chinese & {\cellcolor[HTML]{FFF7FB}} \color[HTML]{000000} 0.1511 & {\cellcolor[HTML]{FFF7FB}} \color[HTML]{000000} 0.1540 & {\cellcolor[HTML]{FFF7FB}} \color[HTML]{000000} 0.1920 & {\cellcolor[HTML]{FFF7FB}} \color[HTML]{000000} 0.1920 & {\cellcolor[HTML]{FFF7FB}} \color[HTML]{000000} 0.2320 & {\cellcolor[HTML]{FFF7FB}} \color[HTML]{000000} 0.1420 & {\cellcolor[HTML]{FFF7FB}} \color[HTML]{000000} 0.2180 & {\cellcolor[HTML]{FFF7FB}} \color[HTML]{000000} 0.1720 \\
P & Detection-Longformer & {\cellcolor[HTML]{FFF7FB}} \color[HTML]{000000} 0.0734 & {\cellcolor[HTML]{FFF7FB}} \color[HTML]{000000} 0.0740 & {\cellcolor[HTML]{FFF7FB}} \color[HTML]{000000} 0.0440 & {\cellcolor[HTML]{FFF7FB}} \color[HTML]{000000} 0.1540 & {\cellcolor[HTML]{FFF7FB}} \color[HTML]{000000} 0.0680 & {\cellcolor[HTML]{FFF7FB}} \color[HTML]{000000} 0.0680 & {\cellcolor[HTML]{FFF7FB}} \color[HTML]{000000} 0.1220 & {\cellcolor[HTML]{FFF7FB}} \color[HTML]{000000} 0.0820 \\
\hline
\end{tabular}
}
\caption{Per-test-language comparison of performance (TPR @ 5\% FPR) of categories of MGT detectors (S -- statistical, P -- pretrained, F -- finetuned). For readability, the finetuned category includes only the best performing combination of train languages of each base model. Bold represents the highest value per each test language.}
\label{tab:categories_tpr}
\end{table*}

\begin{table*}[!t]
\centering
\resizebox{\textwidth}{!}{
\addtolength{\tabcolsep}{-2pt}
\begin{tabular}{c|l||c|c|c|c|c|c|c|c}
\hline
\textbf{Detector} & \textbf{Subset} & \bfseries All & \bfseries cs & \bfseries de & \bfseries hr & \bfseries hu & \bfseries pl & \bfseries sk & \bfseries sl \\
\hline
\multirow[c]{3}{*}{Llama-3.2-3B} & original & {\cellcolor[HTML]{A4BCDA}} \color[HTML]{000000} 0.8821 & {\cellcolor[HTML]{94B6D7}} \color[HTML]{000000} 0.9200 & {\cellcolor[HTML]{A2BCDA}} \color[HTML]{000000} 0.8850 & {\cellcolor[HTML]{86B0D3}} \color[HTML]{000000} 0.9550 & {\cellcolor[HTML]{88B1D4}} \color[HTML]{000000} 0.9500 & {\cellcolor[HTML]{8FB4D6}} \color[HTML]{000000} 0.9300 & {\cellcolor[HTML]{B4C4DF}} \color[HTML]{000000} 0.8350 & {\cellcolor[HTML]{B1C2DE}} \color[HTML]{000000} 0.8400 \\
 & paraphrased & \bfseries \bfseries {\cellcolor[HTML]{89B1D4}} \color[HTML]{000000} 0.9464 & \bfseries \bfseries {\cellcolor[HTML]{80AED2}} \color[HTML]{000000} 0.9700 & \bfseries {\cellcolor[HTML]{8CB3D5}} \color[HTML]{000000} 0.9400 & \bfseries \bfseries {\cellcolor[HTML]{80AED2}} \color[HTML]{000000} 0.9700 & \bfseries \bfseries {\cellcolor[HTML]{80AED2}} \color[HTML]{000000} 0.9700 & \bfseries \bfseries {\cellcolor[HTML]{86B0D3}} \color[HTML]{000000} 0.9550 & \bfseries {\cellcolor[HTML]{8CB3D5}} \color[HTML]{000000} 0.9400 & \bfseries \bfseries {\cellcolor[HTML]{94B6D7}} \color[HTML]{000000} 0.9200 \\
 & homoglyph & {\cellcolor[HTML]{EAE6F1}} \color[HTML]{000000} 0.6357 & {\cellcolor[HTML]{BDC8E1}} \color[HTML]{000000} 0.8050 & {\cellcolor[HTML]{FAF3F9}} \color[HTML]{000000} 0.5350 & {\cellcolor[HTML]{A5BDDB}} \color[HTML]{000000} 0.8750 & {\cellcolor[HTML]{D0D1E6}} \color[HTML]{000000} 0.7500 & {\cellcolor[HTML]{D7D6E9}} \color[HTML]{000000} 0.7200 & {\cellcolor[HTML]{F2ECF5}} \color[HTML]{000000} 0.5850 & {\cellcolor[HTML]{FAF3F9}} \color[HTML]{000000} 0.5350 \\
\hline
\multirow[c]{3}{*}{mDeBERTa-v3-base} & original & {\cellcolor[HTML]{AFC1DD}} \color[HTML]{000000} 0.8500 & {\cellcolor[HTML]{99B8D8}} \color[HTML]{000000} 0.9100 & {\cellcolor[HTML]{B1C2DE}} \color[HTML]{000000} 0.8400 & {\cellcolor[HTML]{CDD0E5}} \color[HTML]{000000} 0.7600 & {\cellcolor[HTML]{8BB2D4}} \color[HTML]{000000} 0.9450 & {\cellcolor[HTML]{B5C4DF}} \color[HTML]{000000} 0.8300 & {\cellcolor[HTML]{ADC1DD}} \color[HTML]{000000} 0.8550 & {\cellcolor[HTML]{DBDAEB}} \color[HTML]{000000} 0.7000 \\
 & paraphrased & \bfseries {\cellcolor[HTML]{97B7D7}} \color[HTML]{000000} 0.9107 & \bfseries {\cellcolor[HTML]{84B0D3}} \color[HTML]{000000} 0.9600 & \bfseries {\cellcolor[HTML]{9CB9D9}} \color[HTML]{000000} 0.9000 & \bfseries {\cellcolor[HTML]{A8BEDC}} \color[HTML]{000000} 0.8700 & \bfseries {\cellcolor[HTML]{81AED2}} \color[HTML]{000000} 0.9650 & \bfseries {\cellcolor[HTML]{A2BCDA}} \color[HTML]{000000} 0.8850 & \bfseries \bfseries {\cellcolor[HTML]{86B0D3}} \color[HTML]{000000} 0.9550 & \bfseries {\cellcolor[HTML]{C6CCE3}} \color[HTML]{000000} 0.7800 \\
 & homoglyph & {\cellcolor[HTML]{FFF7FB}} \color[HTML]{000000} 0.1936 & {\cellcolor[HTML]{FFF7FB}} \color[HTML]{000000} 0.3800 & {\cellcolor[HTML]{FFF7FB}} \color[HTML]{000000} 0.1050 & {\cellcolor[HTML]{FFF7FB}} \color[HTML]{000000} 0.1950 & {\cellcolor[HTML]{FFF7FB}} \color[HTML]{000000} 0.2400 & {\cellcolor[HTML]{FFF7FB}} \color[HTML]{000000} 0.1400 & {\cellcolor[HTML]{FFF7FB}} \color[HTML]{000000} 0.2600 & {\cellcolor[HTML]{FFF7FB}} \color[HTML]{000000} 0.0300 \\
\hline
\multirow[c]{3}{*}{Gemma-2-2B} & original & {\cellcolor[HTML]{B7C5DF}} \color[HTML]{000000} 0.8243 & {\cellcolor[HTML]{9EBAD9}} \color[HTML]{000000} 0.8950 & {\cellcolor[HTML]{A4BCDA}} \color[HTML]{000000} 0.8800 & {\cellcolor[HTML]{ABBFDC}} \color[HTML]{000000} 0.8600 & {\cellcolor[HTML]{8FB4D6}} \color[HTML]{000000} 0.9300 & \bfseries {\cellcolor[HTML]{99B8D8}} \color[HTML]{000000} 0.9100 & {\cellcolor[HTML]{BBC7E0}} \color[HTML]{000000} 0.8150 & {\cellcolor[HTML]{D5D5E8}} \color[HTML]{000000} 0.7300 \\
 & paraphrased & \bfseries {\cellcolor[HTML]{9AB8D8}} \color[HTML]{000000} 0.9029 & \bfseries {\cellcolor[HTML]{84B0D3}} \color[HTML]{000000} 0.9600 & \bfseries \bfseries {\cellcolor[HTML]{8BB2D4}} \color[HTML]{000000} 0.9450 & \bfseries {\cellcolor[HTML]{88B1D4}} \color[HTML]{000000} 0.9500 & \bfseries {\cellcolor[HTML]{81AED2}} \color[HTML]{000000} 0.9650 & \bfseries {\cellcolor[HTML]{99B8D8}} \color[HTML]{000000} 0.9100 & \bfseries {\cellcolor[HTML]{9CB9D9}} \color[HTML]{000000} 0.9000 & \bfseries {\cellcolor[HTML]{B0C2DE}} \color[HTML]{000000} 0.8450 \\
 & homoglyph & {\cellcolor[HTML]{F9F2F8}} \color[HTML]{000000} 0.5414 & {\cellcolor[HTML]{E2DFEE}} \color[HTML]{000000} 0.6700 & {\cellcolor[HTML]{E5E1EF}} \color[HTML]{000000} 0.6600 & {\cellcolor[HTML]{EDE8F3}} \color[HTML]{000000} 0.6200 & {\cellcolor[HTML]{DDDBEC}} \color[HTML]{000000} 0.6950 & {\cellcolor[HTML]{F5EFF6}} \color[HTML]{000000} 0.5650 & {\cellcolor[HTML]{FFF7FB}} \color[HTML]{000000} 0.4250 & {\cellcolor[HTML]{FFF7FB}} \color[HTML]{000000} 0.3400 \\
\hline
\multirow[c]{3}{*}{XLM-RoBERTa-base} & original & {\cellcolor[HTML]{D0D1E6}} \color[HTML]{000000} 0.7529 & {\cellcolor[HTML]{ADC1DD}} \color[HTML]{000000} 0.8550 & {\cellcolor[HTML]{E1DFED}} \color[HTML]{000000} 0.6750 & {\cellcolor[HTML]{AFC1DD}} \color[HTML]{000000} 0.8500 & {\cellcolor[HTML]{A4BCDA}} \color[HTML]{000000} 0.8800 & {\cellcolor[HTML]{C9CEE4}} \color[HTML]{000000} 0.7700 & {\cellcolor[HTML]{C6CCE3}} \color[HTML]{000000} 0.7800 & {\cellcolor[HTML]{E3E0EE}} \color[HTML]{000000} 0.6650 \\
 & paraphrased & \bfseries {\cellcolor[HTML]{BCC7E1}} \color[HTML]{000000} 0.8100 & \bfseries {\cellcolor[HTML]{A2BCDA}} \color[HTML]{000000} 0.8850 & \bfseries {\cellcolor[HTML]{D9D8EA}} \color[HTML]{000000} 0.7100 & \bfseries {\cellcolor[HTML]{A1BBDA}} \color[HTML]{000000} 0.8900 & \bfseries {\cellcolor[HTML]{8EB3D5}} \color[HTML]{000000} 0.9350 & \bfseries {\cellcolor[HTML]{C1CAE2}} \color[HTML]{000000} 0.7950 & \bfseries {\cellcolor[HTML]{A9BFDC}} \color[HTML]{000000} 0.8650 & \bfseries {\cellcolor[HTML]{D6D6E9}} \color[HTML]{000000} 0.7250 \\
 & homoglyph & {\cellcolor[HTML]{FFF7FB}} \color[HTML]{000000} 0.0821 & {\cellcolor[HTML]{FFF7FB}} \color[HTML]{000000} 0.1750 & {\cellcolor[HTML]{FFF7FB}} \color[HTML]{000000} 0.0250 & {\cellcolor[HTML]{FFF7FB}} \color[HTML]{000000} 0.1100 & {\cellcolor[HTML]{FFF7FB}} \color[HTML]{000000} 0.1650 & {\cellcolor[HTML]{FFF7FB}} \color[HTML]{000000} 0.0800 & {\cellcolor[HTML]{FFF7FB}} \color[HTML]{000000} 0.0650 & {\cellcolor[HTML]{FFF7FB}} \color[HTML]{000000} 0.0550 \\
\hline
\multirow[c]{3}{*}{Fast-DetectGPT} & original & \bfseries {\cellcolor[HTML]{FFF7FB}} \color[HTML]{000000} 0.4400 & \bfseries {\cellcolor[HTML]{FFF7FB}} \color[HTML]{000000} 0.4550 & \bfseries {\cellcolor[HTML]{FAF3F9}} \color[HTML]{000000} 0.5350 & \bfseries {\cellcolor[HTML]{FFF7FB}} \color[HTML]{000000} 0.4900 & \bfseries {\cellcolor[HTML]{FFF7FB}} \color[HTML]{000000} 0.4850 & \bfseries {\cellcolor[HTML]{FFF7FB}} \color[HTML]{000000} 0.4600 & \bfseries {\cellcolor[HTML]{FFF7FB}} \color[HTML]{000000} 0.3450 & \bfseries {\cellcolor[HTML]{FFF7FB}} \color[HTML]{000000} 0.4000 \\
 & paraphrased & {\cellcolor[HTML]{FFF7FB}} \color[HTML]{000000} 0.2464 & {\cellcolor[HTML]{FFF7FB}} \color[HTML]{000000} 0.2900 & {\cellcolor[HTML]{FFF7FB}} \color[HTML]{000000} 0.3600 & {\cellcolor[HTML]{FFF7FB}} \color[HTML]{000000} 0.2950 & {\cellcolor[HTML]{FFF7FB}} \color[HTML]{000000} 0.1300 & {\cellcolor[HTML]{FFF7FB}} \color[HTML]{000000} 0.1900 & {\cellcolor[HTML]{FFF7FB}} \color[HTML]{000000} 0.2650 & {\cellcolor[HTML]{FFF7FB}} \color[HTML]{000000} 0.2550 \\
 & homoglyph & {\cellcolor[HTML]{FFF7FB}} \color[HTML]{000000} 0.0014 & {\cellcolor[HTML]{FFF7FB}} \color[HTML]{000000} 0.0000 & {\cellcolor[HTML]{FFF7FB}} \color[HTML]{000000} 0.0000 & {\cellcolor[HTML]{FFF7FB}} \color[HTML]{000000} 0.0050 & {\cellcolor[HTML]{FFF7FB}} \color[HTML]{000000} 0.0000 & {\cellcolor[HTML]{FFF7FB}} \color[HTML]{000000} 0.0050 & {\cellcolor[HTML]{FFF7FB}} \color[HTML]{000000} 0.0050 & {\cellcolor[HTML]{FFF7FB}} \color[HTML]{000000} 0.0050 \\
\hline
\multirow[c]{3}{*}{Binoculars} & original & \bfseries {\cellcolor[HTML]{FFF7FB}} \color[HTML]{000000} 0.4307 & \bfseries {\cellcolor[HTML]{FDF5FA}} \color[HTML]{000000} 0.5150 & \bfseries {\cellcolor[HTML]{FFF7FB}} \color[HTML]{000000} 0.4950 & \bfseries {\cellcolor[HTML]{FDF5FA}} \color[HTML]{000000} 0.5150 & \bfseries {\cellcolor[HTML]{FFF7FB}} \color[HTML]{000000} 0.4900 & \bfseries {\cellcolor[HTML]{FEF6FA}} \color[HTML]{000000} 0.5100 & \bfseries {\cellcolor[HTML]{FFF7FB}} \color[HTML]{000000} 0.3650 & \bfseries {\cellcolor[HTML]{FFF7FB}} \color[HTML]{000000} 0.3900 \\
 & paraphrased & {\cellcolor[HTML]{FFF7FB}} \color[HTML]{000000} 0.2600 & {\cellcolor[HTML]{FFF7FB}} \color[HTML]{000000} 0.3600 & {\cellcolor[HTML]{FFF7FB}} \color[HTML]{000000} 0.3600 & {\cellcolor[HTML]{FFF7FB}} \color[HTML]{000000} 0.3100 & {\cellcolor[HTML]{FFF7FB}} \color[HTML]{000000} 0.1600 & {\cellcolor[HTML]{FFF7FB}} \color[HTML]{000000} 0.2400 & {\cellcolor[HTML]{FFF7FB}} \color[HTML]{000000} 0.3150 & {\cellcolor[HTML]{FFF7FB}} \color[HTML]{000000} 0.2450 \\
 & homoglyph & {\cellcolor[HTML]{FFF7FB}} \color[HTML]{000000} 0.0036 & {\cellcolor[HTML]{FFF7FB}} \color[HTML]{000000} 0.0050 & {\cellcolor[HTML]{FFF7FB}} \color[HTML]{000000} 0.0000 & {\cellcolor[HTML]{FFF7FB}} \color[HTML]{000000} 0.0050 & {\cellcolor[HTML]{FFF7FB}} \color[HTML]{000000} 0.0000 & {\cellcolor[HTML]{FFF7FB}} \color[HTML]{000000} 0.0100 & {\cellcolor[HTML]{FFF7FB}} \color[HTML]{000000} 0.0050 & {\cellcolor[HTML]{FFF7FB}} \color[HTML]{000000} 0.0050 \\
\hline
\multirow[c]{3}{*}{LLM-Deviation} & original & \bfseries {\cellcolor[HTML]{FFF7FB}} \color[HTML]{000000} 0.2071 & \bfseries {\cellcolor[HTML]{FFF7FB}} \color[HTML]{000000} 0.3550 & \bfseries {\cellcolor[HTML]{FFF7FB}} \color[HTML]{000000} 0.2850 & \bfseries {\cellcolor[HTML]{FFF7FB}} \color[HTML]{000000} 0.4400 & \bfseries {\cellcolor[HTML]{FFF7FB}} \color[HTML]{000000} 0.2650 & \bfseries {\cellcolor[HTML]{FFF7FB}} \color[HTML]{000000} 0.3700 & \bfseries {\cellcolor[HTML]{FFF7FB}} \color[HTML]{000000} 0.2250 & \bfseries {\cellcolor[HTML]{FFF7FB}} \color[HTML]{000000} 0.4050 \\
 & paraphrased & {\cellcolor[HTML]{FFF7FB}} \color[HTML]{000000} 0.0943 & {\cellcolor[HTML]{FFF7FB}} \color[HTML]{000000} 0.2450 & {\cellcolor[HTML]{FFF7FB}} \color[HTML]{000000} 0.1450 & {\cellcolor[HTML]{FFF7FB}} \color[HTML]{000000} 0.3600 & {\cellcolor[HTML]{FFF7FB}} \color[HTML]{000000} 0.0350 & {\cellcolor[HTML]{FFF7FB}} \color[HTML]{000000} 0.1600 & {\cellcolor[HTML]{FFF7FB}} \color[HTML]{000000} 0.2000 & {\cellcolor[HTML]{FFF7FB}} \color[HTML]{000000} 0.3050 \\
 & homoglyph & {\cellcolor[HTML]{FFF7FB}} \color[HTML]{000000} 0.0036 & {\cellcolor[HTML]{FFF7FB}} \color[HTML]{000000} 0.0000 & {\cellcolor[HTML]{FFF7FB}} \color[HTML]{000000} 0.0000 & {\cellcolor[HTML]{FFF7FB}} \color[HTML]{000000} 0.0150 & {\cellcolor[HTML]{FFF7FB}} \color[HTML]{000000} 0.0000 & {\cellcolor[HTML]{FFF7FB}} \color[HTML]{000000} 0.0000 & {\cellcolor[HTML]{FFF7FB}} \color[HTML]{000000} 0.0100 & {\cellcolor[HTML]{FFF7FB}} \color[HTML]{000000} 0.0100 \\
\hline
\multirow[c]{3}{*}{BLOOMz-3B-mixed-detector} & original & {\cellcolor[HTML]{FFF7FB}} \color[HTML]{000000} 0.1657 & {\cellcolor[HTML]{FFF7FB}} \color[HTML]{000000} 0.1850 & {\cellcolor[HTML]{FFF7FB}} \color[HTML]{000000} 0.1600 & {\cellcolor[HTML]{FFF7FB}} \color[HTML]{000000} 0.2250 & {\cellcolor[HTML]{FFF7FB}} \color[HTML]{000000} 0.2200 & {\cellcolor[HTML]{FFF7FB}} \color[HTML]{000000} 0.1300 & {\cellcolor[HTML]{FFF7FB}} \color[HTML]{000000} 0.2150 & {\cellcolor[HTML]{FFF7FB}} \color[HTML]{000000} 0.1250 \\
 & paraphrased & \bfseries {\cellcolor[HTML]{FFF7FB}} \color[HTML]{000000} 0.5036 & \bfseries {\cellcolor[HTML]{FBF4F9}} \color[HTML]{000000} 0.5250 & \bfseries {\cellcolor[HTML]{FFF7FB}} \color[HTML]{000000} 0.4250 & \bfseries {\cellcolor[HTML]{F7F0F7}} \color[HTML]{000000} 0.5550 & \bfseries {\cellcolor[HTML]{FAF3F9}} \color[HTML]{000000} 0.5350 & \bfseries {\cellcolor[HTML]{FDF5FA}} \color[HTML]{000000} 0.5150 & \bfseries {\cellcolor[HTML]{F6EFF7}} \color[HTML]{000000} 0.5600 & \bfseries {\cellcolor[HTML]{FFF7FB}} \color[HTML]{000000} 0.4100 \\
 & homoglyph & {\cellcolor[HTML]{FFF7FB}} \color[HTML]{000000} 0.0379 & {\cellcolor[HTML]{FFF7FB}} \color[HTML]{000000} 0.0500 & {\cellcolor[HTML]{FFF7FB}} \color[HTML]{000000} 0.0200 & {\cellcolor[HTML]{FFF7FB}} \color[HTML]{000000} 0.0500 & {\cellcolor[HTML]{FFF7FB}} \color[HTML]{000000} 0.0650 & {\cellcolor[HTML]{FFF7FB}} \color[HTML]{000000} 0.0500 & {\cellcolor[HTML]{FFF7FB}} \color[HTML]{000000} 0.0600 & {\cellcolor[HTML]{FFF7FB}} \color[HTML]{000000} 0.0050 \\
\hline
\multirow[c]{3}{*}{ChatGPT-detector-RoBERTa-Chinese} & original & {\cellcolor[HTML]{FFF7FB}} \color[HTML]{000000} 0.1479 & \bfseries {\cellcolor[HTML]{FFF7FB}} \color[HTML]{000000} 0.1750 & \bfseries {\cellcolor[HTML]{FFF7FB}} \color[HTML]{000000} 0.1850 & {\cellcolor[HTML]{FFF7FB}} \color[HTML]{000000} 0.2050 & {\cellcolor[HTML]{FFF7FB}} \color[HTML]{000000} 0.1950 & {\cellcolor[HTML]{FFF7FB}} \color[HTML]{000000} 0.1850 & {\cellcolor[HTML]{FFF7FB}} \color[HTML]{000000} 0.2750 & \bfseries {\cellcolor[HTML]{FFF7FB}} \color[HTML]{000000} 0.1900 \\
 & paraphrased & \bfseries {\cellcolor[HTML]{FFF7FB}} \color[HTML]{000000} 0.1607 & {\cellcolor[HTML]{FFF7FB}} \color[HTML]{000000} 0.1300 & {\cellcolor[HTML]{FFF7FB}} \color[HTML]{000000} 0.1300 & \bfseries {\cellcolor[HTML]{FFF7FB}} \color[HTML]{000000} 0.2250 & \bfseries {\cellcolor[HTML]{FFF7FB}} \color[HTML]{000000} 0.2700 & \bfseries {\cellcolor[HTML]{FFF7FB}} \color[HTML]{000000} 0.2000 & \bfseries {\cellcolor[HTML]{FFF7FB}} \color[HTML]{000000} 0.2950 & {\cellcolor[HTML]{FFF7FB}} \color[HTML]{000000} 0.1450 \\
 & homoglyph & {\cellcolor[HTML]{FFF7FB}} \color[HTML]{000000} 0.0986 & {\cellcolor[HTML]{FFF7FB}} \color[HTML]{000000} 0.1200 & {\cellcolor[HTML]{FFF7FB}} \color[HTML]{000000} 0.0300 & {\cellcolor[HTML]{FFF7FB}} \color[HTML]{000000} 0.1600 & {\cellcolor[HTML]{FFF7FB}} \color[HTML]{000000} 0.1550 & {\cellcolor[HTML]{FFF7FB}} \color[HTML]{000000} 0.1250 & {\cellcolor[HTML]{FFF7FB}} \color[HTML]{000000} 0.2100 & {\cellcolor[HTML]{FFF7FB}} \color[HTML]{000000} 0.1400 \\
\hline
\multirow[c]{3}{*}{Detection-Longformer} & original & \bfseries {\cellcolor[HTML]{FFF7FB}} \color[HTML]{000000} 0.0714 & \bfseries {\cellcolor[HTML]{FFF7FB}} \color[HTML]{000000} 0.0550 & \bfseries {\cellcolor[HTML]{FFF7FB}} \color[HTML]{000000} 0.0250 & \bfseries {\cellcolor[HTML]{FFF7FB}} \color[HTML]{000000} 0.1350 & \bfseries {\cellcolor[HTML]{FFF7FB}} \color[HTML]{000000} 0.0550 & \bfseries {\cellcolor[HTML]{FFF7FB}} \color[HTML]{000000} 0.0800 & \bfseries {\cellcolor[HTML]{FFF7FB}} \color[HTML]{000000} 0.0900 & \bfseries {\cellcolor[HTML]{FFF7FB}} \color[HTML]{000000} 0.0600 \\
 & paraphrased & {\cellcolor[HTML]{FFF7FB}} \color[HTML]{000000} 0.0450 & {\cellcolor[HTML]{FFF7FB}} \color[HTML]{000000} 0.0500 & {\cellcolor[HTML]{FFF7FB}} \color[HTML]{000000} 0.0150 & {\cellcolor[HTML]{FFF7FB}} \color[HTML]{000000} 0.0900 & {\cellcolor[HTML]{FFF7FB}} \color[HTML]{000000} 0.0300 & {\cellcolor[HTML]{FFF7FB}} \color[HTML]{000000} 0.0750 & {\cellcolor[HTML]{FFF7FB}} \color[HTML]{000000} 0.0500 & {\cellcolor[HTML]{FFF7FB}} \color[HTML]{000000} 0.0450 \\
 & homoglyph & {\cellcolor[HTML]{FFF7FB}} \color[HTML]{000000} 0.0179 & {\cellcolor[HTML]{FFF7FB}} \color[HTML]{000000} 0.0350 & {\cellcolor[HTML]{FFF7FB}} \color[HTML]{000000} 0.0000 & {\cellcolor[HTML]{FFF7FB}} \color[HTML]{000000} 0.0650 & {\cellcolor[HTML]{FFF7FB}} \color[HTML]{000000} 0.0000 & {\cellcolor[HTML]{FFF7FB}} \color[HTML]{000000} 0.0350 & {\cellcolor[HTML]{FFF7FB}} \color[HTML]{000000} 0.0150 & {\cellcolor[HTML]{FFF7FB}} \color[HTML]{000000} 0.0200 \\
\hline
\end{tabular}
}
\caption{Per-test-language comparison of performance (TPR @ 5\% FPR) of the selected MGT detectors on original and adversarial data. Bold represents the highest value per each test language and each detector (in regard to original, paraphrased, and homoglyph subset).}
\label{tab:adversarial_tpr}
\end{table*}

\end{document}